\documentclass{style/nseJournal}

\usepackage[square,sort,comma,numbers]{natbib}

\usepackage{lineno,hyperref}
\usepackage{natbib}
\usepackage{geometry}
\usepackage{fleqn}
\usepackage{graphicx}
\usepackage{newtxtext,newtxmath}
\usepackage{hyperref}
\usepackage{booktabs}

\DeclareMathOperator*{\argmaxA}{arg\,max}
\usepackage{bm}
\usepackage{amsmath}
\usepackage[justification=centering]{caption}
\modulolinenumbers[5]

\usepackage{framed} 
\usepackage{nomencl} 
\makenomenclature

\usepackage{color,soul}
\graphicspath{{./figures/}}

\begin{document}

\title{Quantification of Deep Neural Network Prediction Uncertainties for VVUQ of Machine Learning Models}

\addAuthor{Mahmoud Yaseen}{a}
\addAuthor{\correspondingAuthor{Xu Wu}}{a}
\correspondingEmail{xwu27@ncsu.edu}

\addAffiliation{a}{Department of Nuclear Engineering, North Carolina State University    \\ 
	Burlington Engineering Laboratories, 2500 Stinson Drive, Raleigh, NC 27695 \\}

\addKeyword{Uncertainty Quantification}
\addKeyword{Deep Neural Network}
\addKeyword{Monte Carlo Dropout}
\addKeyword{Deep Ensemble}
\addKeyword{Bayesian Neural Network}

\titlePage

\begin{abstract}
	Recent performance breakthroughs in Artificial intelligence (AI) and Machine learning (ML), especially advances in Deep learning (DL), the availability of powerful, easy-to-use ML libraries (e.g., scikit-learn, TensorFlow, PyTorch.), and increasing computational power have led to unprecedented interest in AI/ML among nuclear engineers. For physics-based computational models, Verification, Validation and Uncertainty Quantification (VVUQ) have been very widely investigated and a lot of methodologies have been developed. However, VVUQ of ML models has been relatively less studied, especially in nuclear engineering. In this work, we focus on UQ of ML models as a preliminary step of ML VVUQ, more specifically, Deep Neural Networks (DNNs) because they are the most widely used supervised ML algorithm for both regression and classification tasks. This work aims at quantifying the prediction, or approximation uncertainties of DNNs when they are used as surrogate models for expensive physical models. Three techniques for UQ of DNNs are compared, namely Monte Carlo Dropout (MCD), Deep Ensembles (DE) and Bayesian Neural Networks (BNNs). Two nuclear engineering examples are used to benchmark these methods, (1) time-dependent fission gas release data using the Bison code, and (2) void fraction simulation based on the BFBT benchmark using the TRACE code. It was found that the three methods typically require different DNN architectures and hyperparameters to optimize their performance. The UQ results also depend on the amount of training data available and the nature of the data. Overall, all these three methods can provide reasonable estimations of the approximation uncertainties. The uncertainties are generally smaller when the mean predictions are close to the test data, while the BNN methods usually produce larger uncertainties than MCD and DE.
\end{abstract}

\section{Introduction}
\label{section:Introduction}

\subsection{Background and Motivation}

Artificial intelligence (AI) aims to build smart machines to automate intellectual tasks normally performed by humans, in a way that we consider intelligent. Machine learning (ML) is a subset of AI that studies computer algorithms which improve automatically through experience (data). ML algorithms typically build a mathematical model based on training data and then make predictions without being explicitly programmed to do so. Its performance increases with experience, in other words, the machine learns. Deep learning (DL) is a subset of ML that uses multi-layered artificial neural networks (ANNs), also called deep neural networks (DNNs), that can automatically learn representations from data without introducing hand-coded rules or human domain knowledge. This is called automatic feature extraction. DNNs' highly flexible architectures can learn directly from large-scale raw data and can increase their predictive accuracy when provided with more data. AI/ML/DL have achieved tremendous success in tasks such as computer vision, natural language processing, speech recognition, and audio synthesis, where the datasets are in the format of images, text, spoken words and videos. Meanwhile, their applications in engineering disciplines mostly focus on scientific data, which resulted in a burgeoning discipline called scientific machine learning (SciML) \cite{baker2019workshop} that blends scientific computing and ML. Typical examples for SciML are data-driven modeling \cite{solomatine2009data} and digital twins \cite{kapteyn2020physics} which have obtained significant interest in the nuclear engineering area in the last few years.

Recent performance breakthroughs in AI and ML, especially advances in DL, the availability of powerful, easy-to-use ML libraries (e.g., scikit-learn, TensorFlow, PyTorch.), and increasing computational power have led to unprecedented interest in AI/ML among nuclear engineers. In the past decade, ML has been leveraged to tackle a diverse range of tasks in nuclear engineering. For example, data-driven closure models in nuclear thermal-hydraulics \cite{liu2019validation} \cite{zhao2020prediction} \cite{bao2021deep}, data-driven material discovery and qualification in nuclear materials \cite{aguiar2020bringing}, autonomous operation and control for advanced reactors \cite{lee2021development} \cite{lin2021development}, data-driven diagnosis and prognosis \cite{lin2022digital},  nuclear data evaluation \cite{vicente2021nuclear}, reactor modeling \& simulation and data-driven surrogates for high-fidelity physics simulations \cite{shriver2021prediction},  nuclear codes calibration, validation and uncertainty quantification \cite{wu2018inversePart1} \cite{wu2018inversePart2} \cite{xie2021towards} \cite{moloko2022quantification}. Such new innovations in ML are beginning to drive advances in nuclear engineering, but the full potential of these techniques for data-driven SciML has yet to be fully realized. One barrier is that existing ML methods often do not meet the needs of nuclear engineering applications. Application-agnostic algorithms, or those designed for more traditional ML applications such as computer vision and natural language processing, cannot typically be directly applied to scientific data in NE and require non-trivial, task-specific modifications. Furthermore, there are significant gaps in the predictive capability assessment and improvement of ML models. Specifically, to enable more trustworthy applications in high-consequence systems like nuclear reactors, the ML practitioners have to consider a few critical unresolved issues, including ML trustworthiness, scaling-induced uncertainty, data scarcity, and lack of Verification, Validation and Uncertainty Quantification (VVUQ) framework for ML applications to evaluate, establish and enhance ML predictive capability/credibility.

\subsection{Prediction/Approximation Uncertainties in ML Models}

For physics-based computational models, VVUQ has been very widely investigated and a lot of methodologies have been developed \cite{oberkampf2010verification}. In brief, \textit{verification} aims to identify, quantify, and reduce errors during the mapping from mathematical model to a computational model, \textit{validation} aims to determine the degree of accuracy of the model in representing real physics. In physics-based modeling and simulation, estimating the uncertainties in output responses is an essential step in model validation and it can establish confidence in the model predictions. UQ methods for physical models provide an estimate of output uncertainties by propagating uncertainties from random input parameters, with methods such as Monte Carlo sampling, stochastic spectral methods and surrogate-based methods, etc. UQ of ML models is equally important, but relatively less studied, especially in nuclear engineering. \textit{The focus of this paper is on UQ of ML models, more specifically, neural networks because they are the most widely used supervised ML algorithm for both regression and classification tasks}.

ML-based models are subject to approximation uncertainties when they are used to make predictions. Such prediction uncertainty is expected to be larger when the training data is limited and predictions are made in extrapolated regions. Generally, noises in data, incomplete coverage of the domain, and imperfect models are the three main sources of uncertainties in ML \cite{abdar2021review} \cite{psaros2022uncertainty}. In the ML community, \textit{epistemic uncertainty} is also called \textit{model uncertainty}, which refers to the uncertainty that results from a lack of training data in certain areas of the input domain. DNNs can perform well at interpolation tasks, but cannot extrapolate or generalize. For example, a DNN trained with cats/dogs images will have difficulty in classifying a bird. \textit{Aleatory uncertainty}, also called \textit{data uncertainty} in ML, is caused by the potential intrinsic randomness of the real data generating process. For example, when predicting the trajectory of an arrow, regression based on archer's force and direction is challenged by randomness in the wind, air pressure and imperfections in the arrow, etc. A combination of various sources of uncertainties can greatly affect the accuracy of DNNs.

\subsection{UQ of DNNs for Nuclear Reactor Simulations}

Note that we would like to point out some inconsistencies in the definitions of uncertainties. In nuclear engineering, and many other disciplines, we categorize uncertainties into two types, \textit{aleatory uncertainties} and \textit{epistemic uncertainties}. Aleatory uncertainties are those arise from randomness (stochastic variations in the physical system). They are irreducible, in the sense that better data or improved models cannot reduce them and can be represented by continuous/discrete random variables. Epistemic uncertainties are due to lack of knowledge, which are considered reducible by more data or information. They are typically represented by continuous/discrete intervals, discrete sets, etc.

We also categorize uncertainties due to their sources \cite{wu2021comprehensive}, as (1) \textit{parameter uncertainty}, originated from ignorance of the exact values of input parameters; (2) \textit{model uncertainty}, due to inaccurate and/or incomplete underlying physics incorporated in the computer models, (3) \textit{numerical uncertainty}, due to numerical approximation errors that cannot be eliminated by the verification process; (4) \textit{experiment uncertainty}, caused by measurement error, and (5) \textit{code/interpolation uncertainty}, owing to the emulation of computationally prohibitive codes using surrogate models (also called metamodels). This source of uncertainty become zero when the original computational model is used instead of a surrogate model. Clearly, the ``sources'' and ``types'' of uncertainties are treated as different concepts. However, we have noticed that in the ML community, many researchers treat ``epistemic uncertainty = model uncertainty'', and ``aleatory uncertainty = data uncertainty'', as described in the previous section. In other words, they have mixed the concepts for sources and types of uncertainties.

\begin{figure}[htb!]
	\centering
	\captionsetup{justification=centering}
	\includegraphics[width=0.6\textwidth]{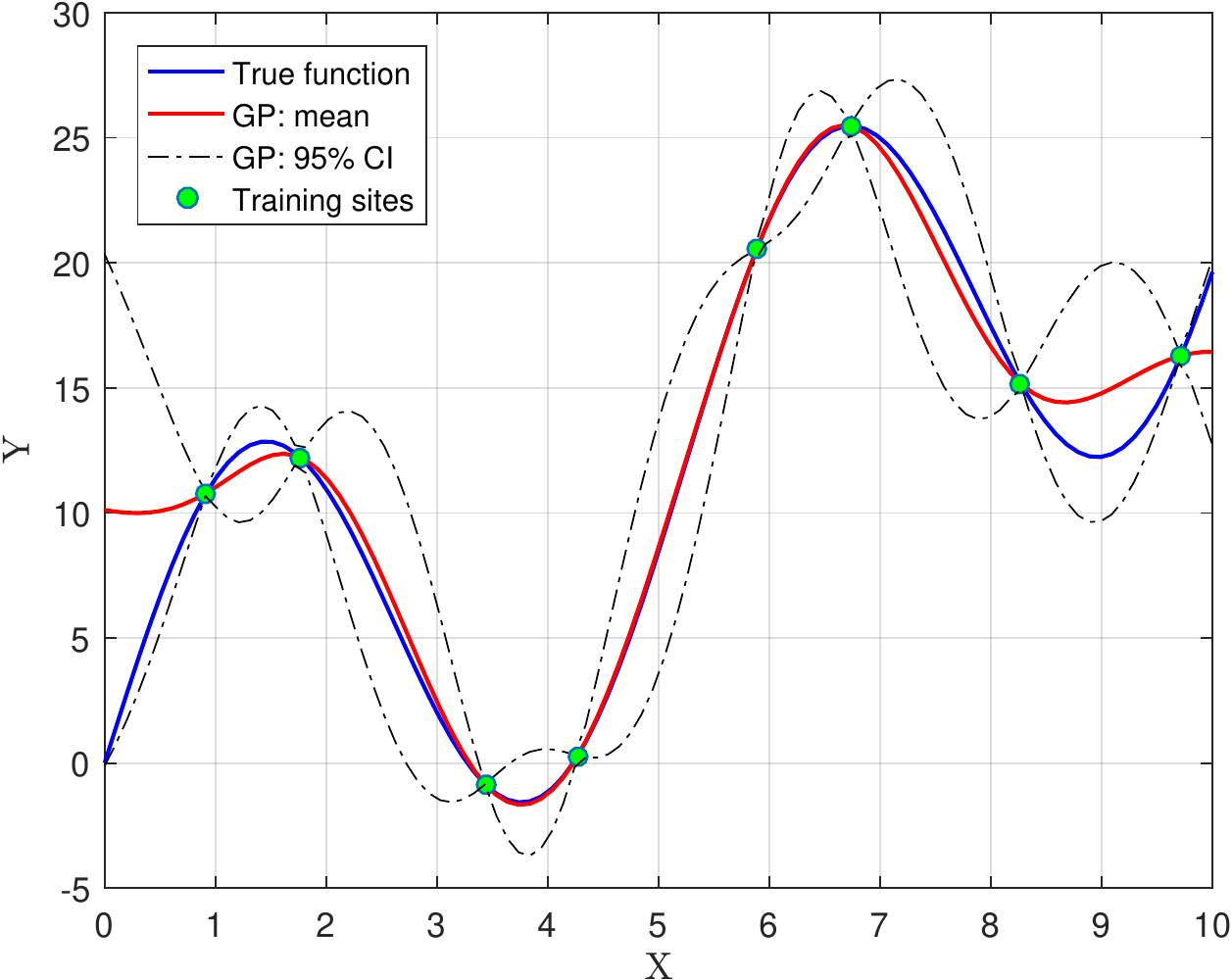}
	\caption[]{Illustration of code/interpolation uncertainty predicted by the GP-based ML models.}
	\label{fig:Intro-GP-Uncertainty}
\end{figure}

In this paper, by ML ``approximation/prediction'' uncertainty, we mean the ``code/interpolation'' uncertainty, for scenarios where the ML models are used as surrogate models for expensive physical models and are trained by the simulation data generated by the physical models. Figure \ref{fig:Intro-GP-Uncertainty} illustrates the code/interpolation uncertainty predicted by Gaussian Process (GP), which is also a widely used algorithm for supervised ML. Based on a few training sites, GP can approximate the true function with a mean function, while at each input, GP's prediction follows a Gaussian distribution. The 95\% confidence interval (CI) indicates the twice of the standard deviation (STD) of the Gaussian distribution. It can be seen that, GP generally interpolates the training sites, and when the input gets far away from any training sites, the GP's prediction uncertainty gets larger, meaning that the GP-based ML model is less confident in its prediction. This is why the code uncertainty is also called interpolation uncertainty. However, GP is the only ML algorithm that can directly estimates such uncertainty, while DNNs cannot. As DNNs are more applicable for a wide range of cases than GP, such as very high dimensional problems, it is necessary to standardize the methods for the UQ of DNNs. Finally, it is worth mentioning that such prediction/approximation uncertainty of DNN, or what we call code/interpolation uncertainty, can be equivalently treated as a combination of ``model (epistemic) + data (aleatory)'' uncertainties used in the ML community. In other words, \textit{it is the total uncertainty introduced when replacing a high-fidelity but computationally prohibitive physics-based model with a ML model (DNN) and use it to make predictions}.

In this paper, we explore and compare three different techniques for quantification of prediction/approximation uncertainties of DNNs, namely Monte Carlo Dropout (MCD) \cite{gal2016dropout}, Deep Ensembles (DE) \cite{lakshminarayanan2017simple} and Bayesian Neural Networks (BNNs) \cite{blundell2015weight}, which have been recently proposed and studied in the ML community but not primarily for scientific data. The uniqueness of this work is that, in the first numerical demonstration example, these three approaches are combined with dimensionality reduction using principal components analysis (PCA) for time-dependent simulations. We use the fission gas release (FGR) \cite{pastore2013physics} model in the Bison fuel performance code developed at Idaho National Laboratory (INL) \cite{williamson2012multidimensional}. DNNs are trained with Bison time series simulation data based on the Risø-AN3 experiment \cite{killeen2006fuel}. MCD, DE and BNN are used in combination with PCA to quantify the uncertainties in the DNN models. This work demonstrates the feasibility of combining supervised ML (regression with DNN) with unsupervised ML (dimensionality reduction with PCA) for UQ of neural network models. In the second example, we applied these approaches to TRACE system thermal-hydraulics (TH) code \cite{USNRC2014TRACE} based on the international OECD/NRC Boiling Water Reactor (BWR) Full-size Fine-Mesh Bundle Tests (BFBT) benchmark \cite{neykov2005nupec}. In this example, TRACE predictions on steady-state void fraction based on BFBT test cases are employed to verify the capabilities of the three UQ approaches.

It was found that the three methods typically require different DNN architectures and hyperparameters to optimize their performance. The UQ results also depend on the amount of training data available and the nature of the data (especially in the Bison example). Overall, all these three methods can provide reasonable estimations of the approximation uncertainties. The uncertainties are generally smaller when the mean predictions are close to the test data, while the BNN method usually produce larger uncertainties than MCD and DE. This paper is organized as follows. Section \ref{section:Introduction} presents the background, motivation, literature review and overview of this manuscript. Section \ref{section:Methodologies} introduces the methodologies for UQ of DNNs. In Section \ref{section:Examples}, details of the numerical demonstration examples are included. Section \ref{section:Results} presents the numerical results and discussions. Section \ref{section:Conclusions} concludes this paper.

\section{Methodologies}
\label{section:Methodologies}

In this section, the MCD, DE and BNN methods are briefly described. With MCD, we will train one DNN for each response, or quantity-of-interest (QoI), and evaluate it multiple times while randomly dropping out some hidden neurons to obtain the prediction uncertainty. With DE, the DNN will directly predict the distributional parameters for the responses (e.g., mean and variance if a Gaussian distribution is assumed). Furthermore, an ensemble of multiple DNNs will be trained for an averaging process. Finally, a BNN assigns uncertain distributions for the DNN parameters (weight and bias), and only one neural network will be trained for each QoI.

\subsection{Monte Carlo Dropout}

Dropout \cite{srivastava2014dropout} is a widely used regularization technique to avoid over-fitting. Unlike $L_1$ and $L_2$ regularizations, dropout doesn't modify the cost function, but the neural network itself. It is applied during the training process of a DNN where certain hidden neurons are randomly ignored or ``dropped out''. This has the effect of making the layer look-like and be treated-like a layer with a different number of nodes and connectivity to the prior layer. Dropout for regularization is used during training but not in prediction, which means, after the training is completed, no hidden neurons will be dropped when making predictions. MCD is a recently developed dropout-based method to quantify DNN prediction uncertainties \cite{gal2016dropout}. The main idea is to use dropout not only during training but also in prediction. The DNN can be evaluated for the same input multiple times, each time with different hidden neurons dropped, resulting in a collection of predictions that can be used to estimate the uncertainties in the prediction.

In a regular feed-forward architecture, the DNN prediction $\hat{y} (\bm{x})$ at input $\bm{x}$ is given by
\begin{equation}
	\hat{y} (\bm{x}) = \sigma ( \ldots \sigma ( \sigma ( \bm{x} \bm{W}_1 + \bm{b}_1 )\bm{W}_2 + \bm{b}_2 ) \ldots  )\bm{W}_L + \bm{b}_L	
\end{equation}
where $L$ is the depth of the neural network, including ($L-1$) hidden layers and 1 output layer, $\sigma$ is the activation function, $\bm{W}_l$ is the matrix of weights for the $l^{\text{th}}$ hidden layer, and $\bm{b}_l$ is the vector of bias for the $l^{\text{th}}$ hidden layer.

With MCD, the DNN prediction is given by
\begin{equation}
	\hat{y} (\bm{x}) = \sqrt{\frac{1}{K_{L-1}}} \sigma ( \ldots \sqrt{\frac{1}{K_2}} \sigma ( \sqrt{\frac{1}{K_1}} \sigma ( \bm{x} \bm{Z}_1 \bm{W}_1 + \bm{b}_1 ) \bm{Z}_2 \bm{W}_2 + \bm{b}_2 ) \ldots  ) \bm{Z}_L \bm{W}_L + \bm{b}_L	
\end{equation}
where $K_l$ is the number of nodes in the $l^{\text{th}}$ hidden layer. $\sqrt{\frac{1}{K_{i}}}$ is added for scaling. $\bm{Z}_l$ is the dropout matrix before the $l^{\text{th}}$ hidden layer with some Bernoulli probability $p_D \in (0, 1)$. The loss function is given by
\begin{equation}
	\mathcal{L}_{\text{MCD}} = \frac{1}{N} \sum_{i=1}^N  \left\Vert \hat{y} (\bm{x}^{(i)}) - y^{(i)} \right\Vert^2  +  \lambda \sum_{l=1}^L \left(  p_D \left\Vert \bm{W}_l \right\Vert_2^2 + \left\Vert \bm{b}_l \right\Vert_2^2  \right)
\end{equation}
where $\lambda$ is the regularization parameter.

The training steps are performed in the regular way, using stochastic gradient descent and re-evaluating the dropout matrices before each learning step. At prediction step, we again evaluate the random dropout matrices before every forward pass, resulting in random neural network outputs. The mean prediction can then be estimated by averaging $T$ forward passes while the uncertainty can be estimated in terms of the empirical variance.
\begin{equation}
	\begin{aligned}    \label{eqn-MCD4-MCD-estimatio-value}
		\hat{\mu} (\bm{x})  &=  \frac{1}{T} \sum_{i=1}^T \hat{y}^{(i)} (\bm{x})    \\
		\hat{\sigma}^2 (\bm{x})  &=  \frac{1}{T - 1} \sum_{i=1}^T \left( \hat{y}^{(i)} (\bm{x}) - \hat{\mu} (\bm{x}) \right)^2
	\end{aligned}
\end{equation}

To sum up, MCD not only randomly turns off hidden neurons for the training step, but also introduce this kind of randomness to the prediction step. If we evaluate the DNN at the same input many times, we can obtain an empirical distribution over the outputs which provides a mean value and a confidence measure in terms of the distributional variance. The empirical variance (i.e., prediction uncertainty) is expected to be low where there is abundant training data since all network subsets have the opportunity to learn in these areas. However, in areas where there is no or little training data to learn from, the DNN behavior is not controllable so we expect a high variance among the different network subsets. In other words, in extrapolation (generalization) regions, the uncertainty will be large, while in the interpolation (training) regions, the uncertainty will be small. This is exactly what we want for uncertainty estimation purposes.

\subsection{Deep Ensemble}

The idea behind DE \cite{lakshminarayanan2017simple} is to train multiple DNN models (called base learners), and then take their average predictions to improve the overall predictive performance. While MCD uses randomness during prediction in order to generate an empirical distribution, DE assumes the data to follow a parameterized distribution where the distribution parameters depend on the input. In other words, the neural network will not output the original QoI, but instead will output the distributional parameters for the QoI. Finding these parameters is the aim of the training process. Without additional knowledge, we usually assume the data to be normally distributed. This leads to very straightforward and easy-to-calculate likelihoods. Because a normal distribution can be fully characterized by its mean $\mu$ and variance $\sigma^2$, the network will try to fit two functions: $\mu_\theta: \mathbb{R}^d \rightarrow \mathbb{R}$ and $\sigma^2_\theta: \mathbb{R}^d \rightarrow (0, +\infty)$, where $\theta$ represents all the network parameters (weights and bias) and $d$ is the dimension of the input features. Because of the significant change in what the DNN predicts directly, we can no longer use the conventional cost/loss functions (such as mean squared error, MSE) since the output is not $y$ anymore, but its mean and variance. Furthermore, the variance $\sigma^2_\theta$ needs to be considered in the new cost function. The Negative Log-Likelihood (NLL) cost function of the normal distribution will be used:
\begin{equation}
	\mathcal{L}_\theta (\bm{x}, y) = - \log p_\theta (y | \bm{x})  =  \frac{\log \hat{\sigma}^2_\theta (\bm{x})}{2}  +  \frac{\left( y - \hat{\mu}_\theta (\bm{x}) \right)^2}{2 \hat{\sigma}^2_\theta (\bm{x})}  +  \text{constant}
\end{equation}
where in the second term, the numerator encourages the mean $\hat{\mu}_\theta (\bm{x})$ to be close to the observed data $y$. The denominator makes sure the variance $\hat{\sigma}^2_\theta (\bm{x})$ is large when the deviation of the mean from the observed data $\left( y - \hat{\mu}_\theta (\bm{x}) \right)^2$ is large. The first term prevents the variance from growing indefinitely. For multiple samples, we take average to use the Mean Negative Log-Likelihood (MNLL):
\begin{equation}
	\mathcal{L}_{\text{MNLL}} = \frac{1}{N} \sum_{i=1}^N \mathcal{L}_\theta \left(  \bm{x}^{(i)}, y^{(i)}  \right)
\end{equation}

For additional robustness, DE uses not only one DNN but an ensemble of multiple DNNs with the same architecture but with different random initializations and training processes. At prediction time, the individual distributions are then averaged for a final estimate, resulting in a Gaussian mixture distribution that can be analyzed for the prediction uncertainty. With an ensemble of $M$ neural networks, the joint Gaussian has mean and variance given by:
\begin{equation}    \label{eqn-DE2-DE-network-mean}
	\hat{\mu}\left(\bm{x}\right) = \frac{1}{M}\sum^{M}_{i = 1} \hat{\mu}_{\theta_i}(\bm{x})
\end{equation}

\begin{equation}    \label{eqn-DE3-DE-network-variance}
	\hat{\sigma}^2 (\bm{x}) = \frac{1}{M} \sum_{i=1}^M \left( \hat{\sigma}^2_{\theta_i} (\bm{x}) + \hat{\mu}^2_{\theta_i} (\bm{x}) \right)  -  \hat{\mu}^2 (\bm{x})
\end{equation}
where $\hat{\mu}\left(\bm{x}\right),  \hat{\sigma}^2\left(\bm{x}\right)$ are the mean and variance of networks ensemble respectively.

\subsection{Bayesian Neural Network}

In a regular DNN, we start from random initialization of the network parameters (weights, bias), then we update them using stochastic gradient descent, in which the derivative of the cost function w.r.t. the parameters are calculated with backpropagation. After training/learning for multiple epochs, we obtain deterministic estimations of the network parameters. When an input enters the trained neural network, we get a deterministic output. A BNN \cite{ghahramani2016history} \cite{goan2020bayesian} is a neural network with distributions over parameters (weights, bias). In BNNs, a prior distribution is first specified upon each of the neural network parameters, then, given the training data, the posterior distributions over the parameters are computed. After training, the BNN can be evaluated at the same input for multiple times, each time with samples for its parameters from the posterior distributions, thus producing different values for the prediction that can be used to quantify the predictive uncertainties.

The greatest challenge in BNN is that after getting an approximation formulation for the posterior distributions of the BNN parameters, it is difficult to solve for these posterior distributions. Since exact Bayesian inference is computationally intractable for neural networks, a variety of approximation methods have been developed including Markov Chain Monte Carlo (MCMC) \cite{neal2012bayesian} and variational Bayesian inference \cite{blei2017variational} \cite{tzikas2008variational}, etc. MCMC is known for slow convergence that is also difficult to diagnose. Variational inference methods have become popular as they shift from sampling to optimization, which makes them much faster than MCMC for complex models and larger datasets. In the following, we will briefly introduce the idea behind solving for BNN with Variational Inference (VI), also referred to as Variational Bayesian (VB). Note that we will treat the bias parameters as deterministic values as they do not affect the neural network predictions as much as the weight parameters. This is because of the way in which the weights and bias are used in the activation function. For example, in the sigmoid activation function, the weight parameters are multiplied by the activation from the previous layer, while the bias parameters are added to the former product. Therefore, the bias parameters affect the DNN prediction less significantly than the weights.

The method use VI for BNN weights training is referred to as ``Bayes by Backprop'' in \cite{blundell2015weight}. To better explain this method, we first interpret a neural network as a probabilistic model: $P (\mathbf{y} | \mathbf{x}, \mathbf{w})$, where $\mathbf{x} \in \mathbb{R}^d$ is the $d$-dimensional input vector, $\mathbf{w}$ is a set of parameters such as weights, and $\mathbf{y}$ is the output. The weights can be learnt by Maximum Likelihood Estimation (MLE): given a set of training data $\mathcal{D} = \left\{ (\mathbf{x}_i, \mathbf{y}_i) \right\}_{i=1}^{N}$, the MLE of the weights $\mathbf{w}^{\text{MLE}}$ are defined as:
\begin{equation}
	\mathbf{w}^{\text{MLE}}  =  \argmaxA_{\mathbf{w}}  \log P (\mathcal{D} | \mathbf{w})   =  \argmaxA_{\mathbf{w}}  \sum_{i=1}^{N} \log P (\mathbf{y}_i | \mathbf{x}_i, \mathbf{w})
\end{equation}

The training process is done by employing gradient descent with backpropagation, where we assume that $\log P (\mathcal{D} | \mathbf{w})$ is differentiable in $\mathbf{w}$. BNN calculates the posterior distribution $P (\mathbf{w} | \mathcal{D})$ of the weights given the training data based on the Bayes' rule:
\begin{equation}
	P (\mathbf{w} | \mathcal{D})  =  \frac{P (\mathcal{D} | \mathbf{w}) \cdot P (\mathbf{w})}{P (\mathcal{D})}
\end{equation}
where $P (\mathbf{w})$ is the prior distribution for $\mathbf{w}$, $P (\mathcal{D} | \mathbf{w})$ is the likelihood function, and $P (\mathbf{w} | \mathcal{D})$ is the posterior distribution for $\mathbf{w}$. Prior and posterior represent our knowledge of $\mathbf{w}$ before and after observing the training data $\mathcal{D}$, respectively. $P (\mathcal{D})$ does not contain $\mathbf{w}$ so it is usually treated as a normalizing constant. It is sometimes referred to as the evidence term.

When making predictions at a test data $\mathbf{x}^{*}$, the predictive distribution of the output $\mathbf{y}^{*}$ is given by:
\begin{equation}
	P (\mathbf{y}^{*} | \mathbf{x}^{*})  =  \mathbb{E}_{P (\mathbf{w} | \mathcal{D})} \left[  P (\mathbf{y}^{*} | \mathbf{x}^{*}, \mathbf{w})  \right]
\end{equation}
where the expectation operator $\mathbb{E}_{P (\mathbf{w} | \mathcal{D})}$ means we need to integrate over $P (\mathbf{w} | \mathcal{D})$. Each possible configuration of the weights, weighted according to the posterior distribution $P (\mathbf{w} | \mathcal{D})$, makes a prediction about $\mathbf{y}^{*}$ given $\mathbf{x}^{*}$. This is why taking an expectation under the posterior distribution on weights is equivalent to using an ensemble of an infinite number of neural networks. Unfortunately, such expectation operation is intractable for neural networks of any practical size, due to large number of parameters as well as the difficulty to perform exact integration.

This is the primary motivation to use a variational approximation for $P (\mathbf{w} | \mathcal{D})$. VI methods are a family of techniques for approximating intractable integrals arising in Bayesian inference and ML. VI is used to approximate complex posterior probabilities that are difficult to evaluate directly as an alternative strategy to MCMC sampling. The general idea is to use a variational distribution $Q (\mathbf{w} | \bm{\theta})$, which is a family of distributions controlled by parameters $\bm{\theta}$, to approximate $P (\mathbf{w} | \mathcal{D})$. For example $\bm{\theta} = \{ \mu, \sigma \}$ if Gaussian distribution is chosen. We seek to find optimal value $\bm{\theta}^{*}$ of $\bm{\theta}$ that can minimize the Kullback-Leibler divergence (KLD) from $P (\mathbf{w} | \mathcal{D})$ to $Q (\mathbf{w} | \bm{\theta})$:
\begin{equation}
	\begin{aligned}
		\bm{\theta}^{*}  
		&=  \argmaxA_{\bm{\theta}}  D_{\text{KL}} \left( Q (\mathbf{w} | \bm{\theta}) \Vert P (\mathbf{w} | \mathcal{D}) \right)   
		=  \argmaxA_{\bm{\theta}}  \int  Q (\mathbf{w} | \bm{\theta}) \log  \frac{Q (\mathbf{w} | \bm{\theta})}{P (\mathbf{w} | \mathcal{D})}   d \mathbf{w}    \\
		&=  \argmaxA_{\bm{\theta}}  \int  Q (\mathbf{w} | \bm{\theta}) \log  \frac{Q (\mathbf{w} | \bm{\theta})}{P (\mathcal{D} | \mathbf{w}) \cdot P (\mathbf{w})}  d \mathbf{w}    \\
		&=  \argmaxA_{\bm{\theta}}  \left(  D_{\text{KL}} \left( Q (\mathbf{w} | \bm{\theta}) \Vert P (\mathbf{w}) \right)  -  \mathbb{E}_{Q (\mathbf{w} | \bm{\theta})} \left[  \log P (\mathcal{D} | \mathbf{w})  \right]  \right)
	\end{aligned}
\end{equation}

The resulting cost/loss function, denoted as $\mathcal{F} (\mathcal{D}, \bm{\theta})$, is known as the \textit{variational free energy}, or \textit{expected lower bound}:
\begin{equation}
	\mathcal{F} (\mathcal{D}, \bm{\theta})  =  D_{\text{KL}} \left( Q (\mathbf{w} | \bm{\theta}) \Vert P (\mathbf{w}) \right)  -  \mathbb{E}_{Q (\mathbf{w} | \bm{\theta})} \left[  \log P (\mathcal{D} | \mathbf{w})  \right]
\end{equation}
where the first part of $\mathcal{F} (\mathcal{D}, \bm{\theta})$ is prior-dependent and referred to as the complexity cost. The second part is data-dependent and referred to as the likelihood cost. The cost function embodies a trade-off between matching the data $\mathcal{D}$ (likelihood) and satisfying the simplicity prior $P (\mathbf{w})$. Further simplifications are needed for $\mathcal{F} (\mathcal{D}, \bm{\theta})$ in order to evaluate it numerically. Interested readers are recommended to look at \cite{blundell2015weight} \cite{blei2017variational} for more details.

\section{Demonstration Examples}
\label{section:Examples}

\subsection{Demonstration Example on Bison}

\subsubsection{Problem Definition and Training Data}

In this example, we consider a realistic nuclear engineering problem to investigate the capabilities of MCD, DE and BNN, which is the physics-based FGR model \cite{pastore2013physics} in the Bison code \cite{williamson2012multidimensional}. Bison is a finite element-based nuclear fuel performance code developed at Idaho National Laboratory (INL). Nuclear reactor fuel performance studies the thermo-mechanical behavior of fuel rods and verify their compliance with safety criteria under both normal operation and accidental conditions. The complex behavior of the fission gases xenon and krypton in UO\textsubscript{2} significantly affects the thermo-mechanical performance of the nuclear fuel rods due to multiple reasons, such as pellet-cladding mechanical interaction caused by fuel swelling and fuel rod gap closure, pressure build-up and thermal conductivity degradation of the fuel rod filling gas (helium). The accurate modeling of fission gas behavior in nuclear fuel performance simulation is vital considering its detrimental nature.

Numerical analysis of FGR and swelling involves treatment of several complicated and interrelated physical processes, which inevitably depend on uncertain input parameters. For example, the state-of-the-art fuel performance code Bison \cite{williamson2012multidimensional} incorporates an advanced physics-based FGR model that depends on  5 model parameters whose uncertainties are only known by expert judgement \cite{pastore2015uncertainty}. All of these  5 uncertain inputs are characterized as multiplicative/scale factors, as shown in Table \ref{table:Bison0-Parameters}. The lower and upper bounds are suggested by the Bison FGR model developers in a uncertainty and sensitivity study \cite{pastore2015uncertainty}. Such uncertainty bounds specification is based on the scatter in the available experimental data from a fairly extensive literature review and is consistent with the information in the open literature. Normal distributions are assumed for the first two parameters. The last three parameters adopt log-uniform distributions so that the logarithms of them follow uniform distribution on $[-1, 1]$. We can see that the uncertainty ranges for the last three parameters have lower and upper bounds that differ by a factor of 100. Apparently a more robust and appropriate specifications of the input uncertainties should be sought that are consistent with available experimental data. Such a need motivates our earlier study \cite{wu2018kriging} on inverse UQ of the uncertainties in Bison FGR physical model parameters, based on Risø-AN3 benchmark which includes on-line, time-dependent measurement of FGR.

\begin{table}[htbp!]
	\small
	\captionsetup{justification=centering}
	\caption{Uncertain input parameters in the Bison FGR model.}
	\label{table:Bison0-Parameters}
	\centering
	\begin{tabular}{l c c c c}
		\toprule
		Parameter (scale factor) & Description & Lower bound & Upper bound & Nominal \\ 
		\midrule
		Temperature 								& \texttt{temperature\_scalef}	& 0.95 & 1.05 & 1.0 \\
		Grain radius 								& \texttt{grainradius\_scalef} 	& 0.4 & 1.6  & 1.0 \\
		Intra-granular gas atom diffusion coeff. 	& \texttt{igdiffcoeff\_scalef} 	& 0.1 & 10.0 & 1.0 \\
		Intra-granular resolution parameter 		& \texttt{resolutionp\_scalef} 	& 0.1 & 10.0 & 1.0 \\
		Grain-boundary diffusion coeff. 			& \texttt{gbdiffcoeff\_scalef} 	& 0.1 & 10.0 & 1.0 \\
		\bottomrule
	\end{tabular}
\end{table}

\begin{figure}[htb!]
	\centering
	\captionsetup{justification=centering}
	\includegraphics[width=0.6\textwidth]{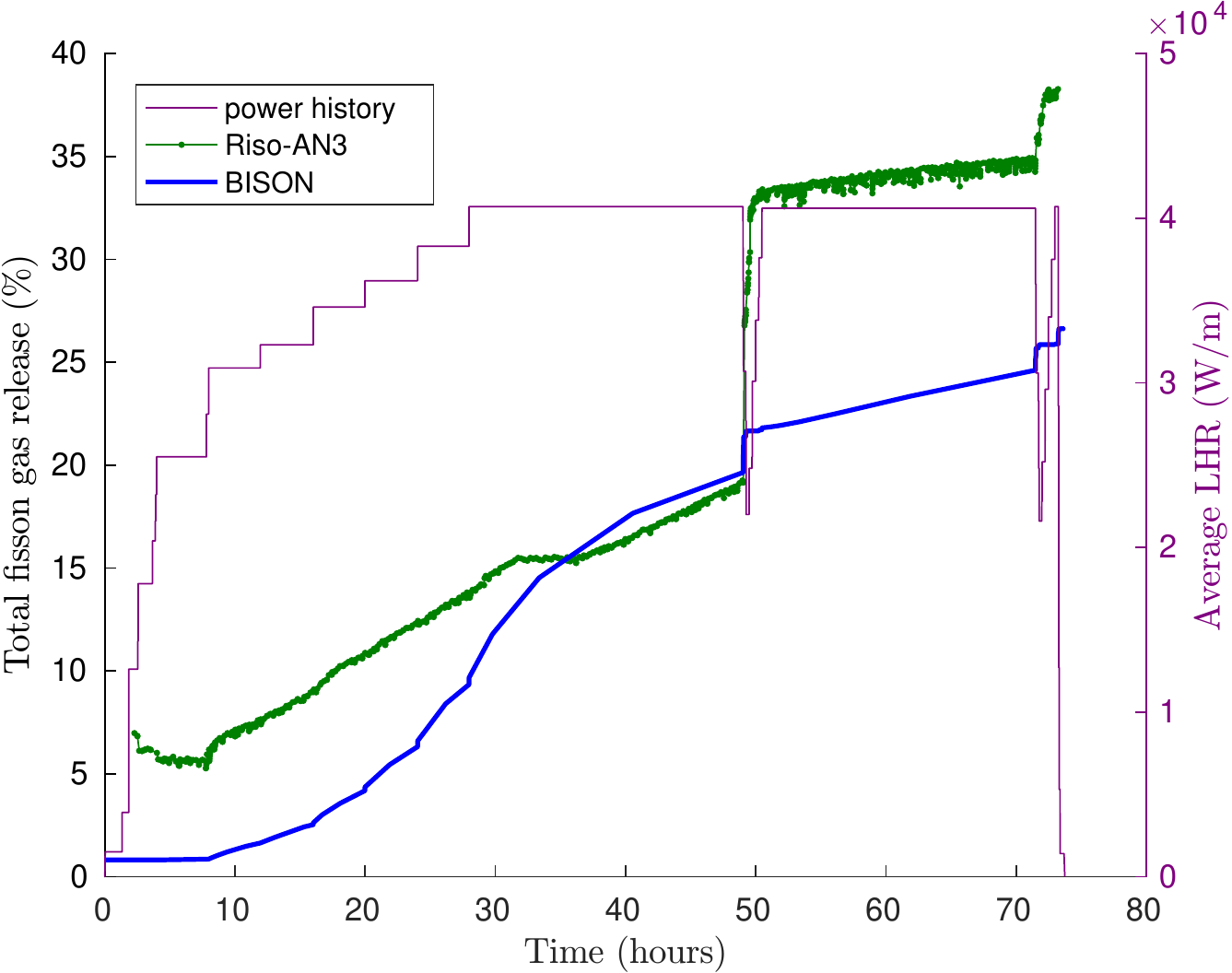}
	\caption[]{Comparison of FGR time series from Risø-AN3 measurement and Bison simulation with nominal values of uncertain input parameters.}
	\label{fig:Bison0-FGR}
\end{figure}

The Risø-AN3 experiment is one of the fuel rod irradiation experiments from the International Fuel Performance Experiments (IFPE) database \cite{killeen2006fuel}. It comprises base irradiation in the BIBLIS A PWR (Germany) and ramp test in the DR3 research reactor at Risø (Denmark). Accurately predicting FGR in fuel performance codes is known to be extremely challenging. Figure \ref{fig:Bison0-FGR} shows the comparison of FGR from Bison simulation at nominal input values of these  5 parameters and Risø-AN3 measurement data during a ramp test. The FGR is defined as the ratio between the amount of fission gas generated and released in the irradiated fuel rod. Bison under-predicts the FGR over the most of the ramp test. There is a burst release of FGR due to the crack opening of the fuel pellets when the power suddenly drops at 50 hours after the start of transient. Bison is able to qualitatively reproduce the rapid increase of FGR around 50 hours, albeit only by a much lower magnitude. A possible explanation, as pointed out in \cite{williamson2016validating}, is that the magnitude of the release burst measured by pressure transducer is over-estimated. Another reason is that the gap and cracking opening during the sudden power drop can cause delayed detection of gas released from the fuel prior to the transient \cite{killeen2006fuel}.

In a previous work \cite{wu2018kriging}, we developed a Bayesian inverse UQ methodology to quantify the parameter uncertainties in these  5 uncertain inputs based on the Risø-AN3 data. Inverse UQ is defined as the process to inversely quantify the input uncertainties based on experimental data. For more details, see a review paper on inverse UQ \cite{wu2021comprehensive}. Bayesian inverse UQ first formulates posterior distributions for the uncertain inputs, then uses MCMC sampling to explore the non-standard posterior distributions. The primary challenge with MCMC is that typically $\sim$10,000 samples are needed, each requiring an evaluation of the computational model which can be computationally prohibitive. In the previous work \cite{wu2018kriging}, GP-based surrogate model was used to replace Bison during MCMC sampling to significantly reduce the computational cost. However, GP is unable to scale well for high-dimensional problems. This is the motivation for us to consider the more flexible DNNs. Compared to DNNs, the major advantage of GP over DNN is that the approximation uncertainties is directly available in GP, as shown in Figure \ref{fig:Intro-GP-Uncertainty}. Therefore, in this work, we seek to demonstrate the capabilities of MCD, DE and BNN in quantifying the DNN prediction uncertainties.

\begin{figure}[htb!]
	\centering
	\captionsetup{justification=centering}
	\includegraphics[width=0.6\textwidth]{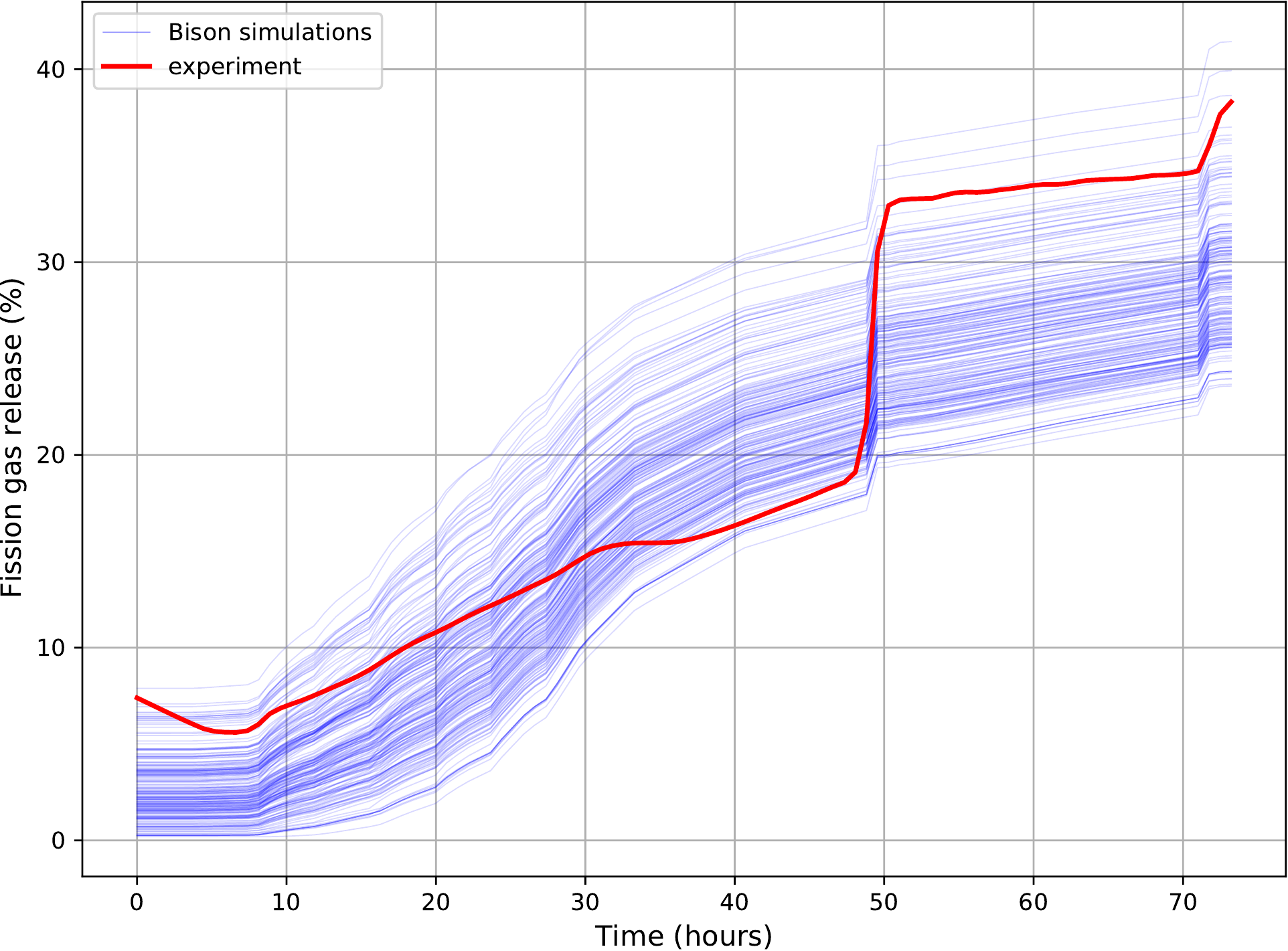}
	\caption[]{Bison time series FGR simulations using 200 samples of inputs.}
	\label{fig:Bison0-DoE}
\end{figure}

To train the DNN-based surrogate model, we generate training data by running Bison at 200 samples of the  5 uncertain inputs, which are obtained from a design of experiment process using maximin Latin hypercube sampling (LHS) with 1,000 iterations. This means we generate 1,000 sample designs with LHS and select the one with the maximum minimum distance such that the input space is sufficiently explored. Based on the previous work, 100 samples were sufficient for the training of GP. Since Bison code is very expensive to run (2.5 hours for 1 simulation with 32 processors), the fact that a surrogate model requires many more samples to train means that it is not an appropriate method. As DNNs are more parameterized than GP (a DNN has a lot more unknown parameters including weights and bias), we use 200 samples in this work and consider it an acceptable number. Figure \ref{fig:Bison0-DoE} shows the Bison simulation results at 200 samples. Note that in this work Risø-AN3 data is not used, because the primary interest in to build a surrogate model for Bison code using DNN and quantify its prediction uncertainties, so only Bison simulations are used as training data.

\subsubsection{Dimensionality Reduction for Transient Training Data}

The major challenge in this demonstration example is that the training dataset consists of 200 Bison simulation curves of FGR. Time-dependent simulations correspond to infinite-dimensional responses. We can select a few points from the Bison simulation curve to represent FGR responses at different times. However, the number of points (hence dimension of the responses) cannot be low in order to sufficiently cover the whole transient. To address this issue, we adopted a similar strategy used in \cite{wu2018kriging} to perform dimensionality reduction of the transient simulation with PCA \cite{shlens2014tutorial}, which is a major tool in unsupervised ML. PCA is the projection of high-dimensional data onto a lower-dimensional subspace (also called principal subspace) such that the projected data has maximized variance and is uncorrelated. PCA can be done through eigen-decomposition of the covariance matrix of the data. However, Singular Value Decomposition (SVD) of the data matrix is often preferred for numerical reasons.

We first pick $p=100$ points from Bison FGR time series simulation, which are equally distributed through the transient but with a few more points at around 50 hours. With $N=200$ simulation curves, the training dataset can be represented by a $p \times N$ data matrix $\mathbf{A}$. The rows correspond to FGR responses at different times, while the columns means different samples. There are high correlations between the rows of this data matrix and we would like to de-correlate the data with a linear transformation $\mathbf{P} \mathbf{A} = \mathbf{B}$, where $\mathbf{P}$ is a $p \times p$ transformation matrix, $\mathbf{B}$ is a $p \times N$ data matrix and the rows of $\mathbf{B}$ represent uncorrelated new responses. PCA will be used to find $\mathbf{P}$.

The first step of PCA is to center the original data matrix $\mathbf{A}$. Define $\mathbf{u}$ as the column vector of the row means. We get the centered data matrix $\mathbf{A}_{\text{c}}$ by subtracting $\mathbf{u}$ from each column of $\mathbf{A}$. The second step is to find the SVD of $\mathbf{A}_{\text{c}}$ with $\mathbf{A}_{\text{c}} = \mathbf{U} \bm{\Lambda} \mathbf{V}^\top$, where $\mathbf{U}$ is a $p \times p$ orthogonal matrix whose columns are the left-singular vectors of $\mathbf{A}_{\text{c}}$, $\mathbf{V}$ is a $N \times N$ orthogonal matrix whose columns are the right-singular vectors of $\mathbf{A}_{\text{c}}$, and $\bm{\Lambda}$ is a $p \times N$ diagonal matrix whose diagonal entries are the singular values of $\mathbf{A}_{\text{c}}$. The non-zero singular values are the square roots of the non-zero eigenvalues of $\mathbf{A}_{\text{c}} \mathbf{A}_{\text{c}}^\top$ or $\mathbf{A}_{\text{c}}^\top \mathbf{A}_{\text{c}}$. The magnitude of the diagonal entries of $\bm{\Lambda}$ are arranged in descending order and large singular values points to important features in data matrix $\mathbf{A}_{\text{c}}$. Next, by choosing $\mathbf{P} = \mathbf{U}^\top$, we have $\mathbf{P} \mathbf{A}_{\text{c}} = \mathbf{U}^\top \mathbf{A}_{\text{c}} = \bm{\Lambda} \mathbf{V}^\top = \mathbf{B}$. It can be seen that the covariance matrix of $\mathbf{B}$ is diagonal. The rows of $\mathbf{P}$ (which are the columns of $\mathbf{U}$) are called the \textit{Principal Components (PCs)}, also known as \textit{loadings}. The transformed responses (rows of $\mathbf{B}$) are called \textbf{PC scores}. PC scores are the representations of the original data $\mathbf{A}_{\text{c}}$ in the principal subspace. The columns of $\mathbf{B}$ correspond to observations. The eigenvalues of the covariance matrix of $\mathbf{A}_{\text{c}}$ are called \textit{PC variances}, which correspond to the square of the diagonal entries in $\bm{\Lambda}$.

The last step is to determine the dimension of the principal subspace $p^{*}$ which is much smaller than $p$. It is quite normal that the PC variances decrease very quickly. PCs with larger associated variances represent important structure, while those with lower variances represent noise. A widely used criterion is that we only keep the first $p^{*}$ PCs whose corresponding PC variances can account for over 95\% or 99\% of the total variance. The first $p^{*}$ PCs form a $p^{*} \times p$ transformation matrix $\mathbf{P}^{*}$:
\begin{equation}
	\mathbf{P}^{*} \mathbf{A}_{\text{c}} = \mathbf{B}^{*}
\end{equation}
where $\mathbf{B}^{*}$ is a $p^{*} \times N$ matrix whose rows represented new responses after dimensionality reduction. Now we have reduced the number of responses from $p$ to $p^{*}$. Once we have the PCs as rows of matrix $\mathbf{P}^{*}$ and PC scores stored in matrix $\mathbf{B}^{*}$, we build the DNN-based surrogate model for the $p^{*}$ PC scores instead of the $p$ output responses in the original space. The data matrix $\mathbf{B}^{*}$ will serve as the training samples. The DNN prediction for a certain input will be a $p^{*} \times 1$ vector $\mathbf{b}^{*}$, where the ``${*}$'' superscript is used to indicate that it is a prediction in PC subspace. We can transformed it back to the original space with $\mathbf{P}^{*}$:
\begin{equation}    \label{eqn-PCA1-reverse-PCA}
	\mathbf{a} = {\mathbf{P}^{*}}^\top \mathbf{b}^{*} + \mathbf{u}
\end{equation}
where is a $p \times 1$ vector represents the FGR time series in the original space. Note that we need to add the mean vector $\mathbf{u}$ back to obtain the FGR data.

\begin{figure}[htb!]
	\centering
	\captionsetup{justification=centering}
	\includegraphics[width=0.6\textwidth]{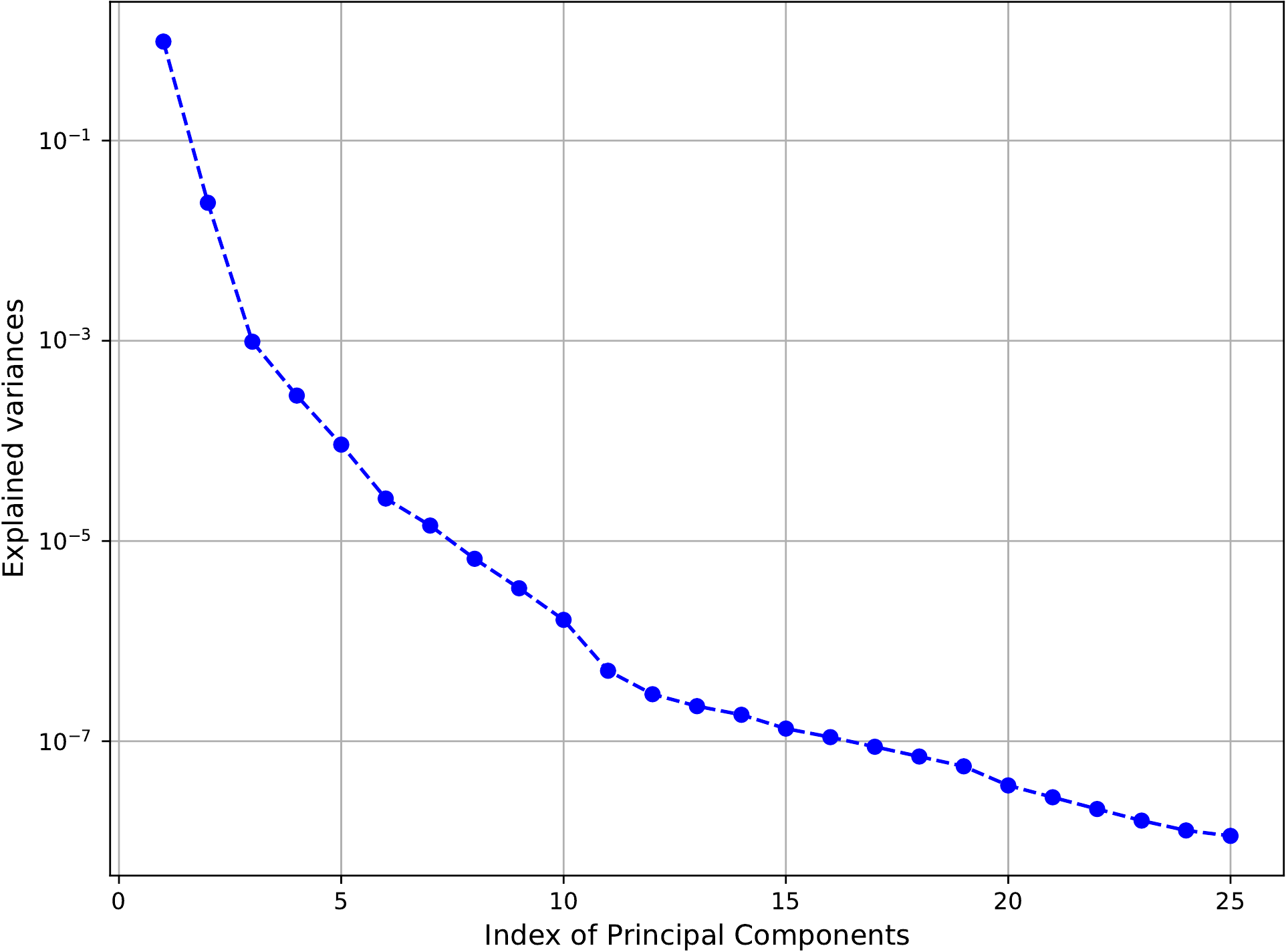}
	\caption[]{Decay of the explained variance of the first 25 PCs.}
	\label{fig:Bison0-PC-Variances}
\end{figure}

\begin{figure}[htb!]
	\centering
	\captionsetup{justification=centering}
	\includegraphics[width=0.6\textwidth]{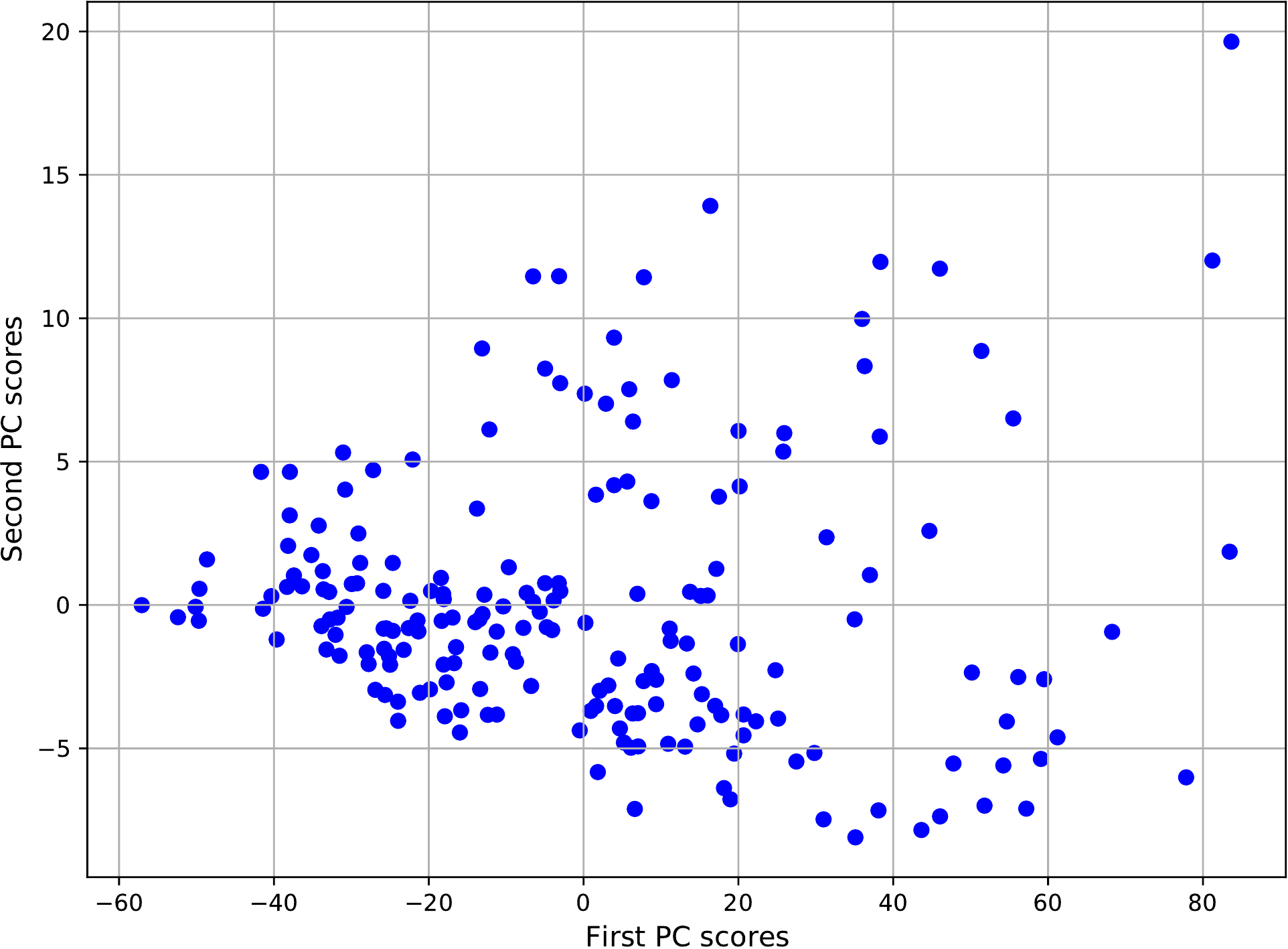}
	\caption[]{PC scores corresponding to the first 2 PCs after dimensionality reduction.}
	\label{fig:Bison0-PC-Scores}
\end{figure}

To find the dimension $p^{*}$ for the PC subspace, we need to determine the number of significant PCs that can explain 99\% of the total variance in the original dataset. Figure \ref{fig:Bison0-PC-Variances} shows the decrease of the PC variances for the first 25 principal dimensions. It can be seen that the PC variances drop so quickly that the variances associated with higher index become trivial compared with the first few variances (note that y axis uses log scale). The percentage of variation explained is the ratio of the PC variance in certain principal dimension over the sum of the variances in all the principal dimensions. The first 2 PC dimensions together are able to account for over 99.5\% of the total variation, suggesting that we only need to keep these 2 dimensions. The new responses are the corresponding PC scores, as shown in Figure \ref{fig:Bison0-PC-Scores}. The original time series FGR curve can be reconstructed from reverse PCA using the loading vectors obtained from (forward) PCA, as shown in Equation (\ref{eqn-PCA1-reverse-PCA}).

To sum up, for the training of DNNs, the inputs are the  5 uncertain parameters in the Bison FGR model, while the outputs are the scores corresponding to the first two PCs of the time-dependent Bison simulation data for FGR. We will employ MCD, DE and BNN to quantify the uncertainties in the DNN predictions of the PC scores, and then project the uncertainties back to the original space to get the uncertainties in the FGR time series data.

\subsection{Demonstration Example on TRACE}

In this example, training data generated by TRACE simulation will be used. The motivation for this example is similar to Bison, which is to build DNN-based surrogate models with quantified uncertainties that can be used during MCMC sampling for Bayesian inverse UQ. However, unlike the Bison example, in which the training data is from BISON simulation of time-dependent FGR based on a single experiment, for this example the training data is generated by TRACE void fraction simulations based on the BFBT benchmark \cite{neykov2005nupec} which consists of many experiments.

For best-estimate TH codes, significant uncertainties also come from the closure laws (also known as correlations or constitutive relationships) which are used to describe the transfer terms in the balance equations. These physical models govern the mass, momentum and energy exchange between the fluid phases and surrounding medium, varying according to the type of two-phase flow regime. When the closure models were originally developed, their accuracy and reliability were studied with a particular experiment. However, once they are implemented in a TH code as empirical correlations and used for prediction of different physical systems, the accuracy and uncertainty characteristics of these correlations are no longer known to the code user. Previously in the uncertainty and sensitivity study of such codes, physical model uncertainties are simply ignored, or described using expert opinion or self-judgment. In a previous study \cite{wu2018inversePart1}, we developed a modular Bayesian analysis method to inversely quantify the TRACE physical model parameter uncertainties, and applied it based on the BFBT data \cite{wu2018inversePart2}. GP models were also used as a surrogate model to TRACE to significantly reduce the computational cost in MCMC sampling. Similarly, in this study, we want to explore the potential of using DNN as a surrogate model for TRACE using the same training dataset, with a focus on quantification of the DNN uncertainties.

The BFBT facility is full-scale BWR assembly, with measurement performed under typical reactor power and high-pressure, high-temperature fluid conditions found in BWRs. Two types of BWR assemblies are simulated in a full length test facility, a current 8$\times$8 fuel bundle and a 8$\times$8 high burn-up bundle, where each rod is electrically heated to simulate an actual reactor fuel rod.  5 test assembly configurations (types 0, 1, 2, 3 and 4) with different geometries and power profiles were utilized for the void distribution and critical power measurements. In this example, we only consider the test assembly 4101, which was also used in our previous study on Bayesian inverse UQ \cite{wu2018inversePart2}.

\begin{figure}[!ht]
	\centering
	\captionsetup{justification=centering}
	\includegraphics[width=0.5\textwidth]{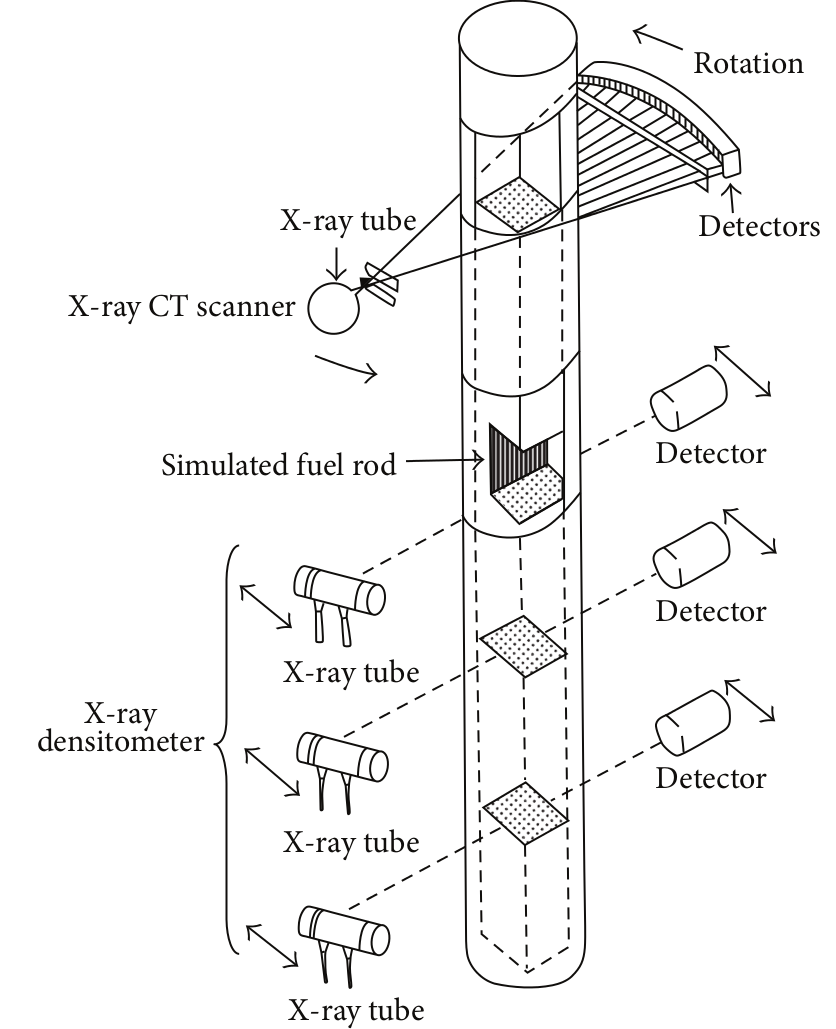}
	\caption[]{BFBT void fraction measurement structure.}
	\label{fig:TRACE0-BFBT}
\end{figure}

Two types of void distribution measurement systems were employed: an X-ray computer tomography (CT) scanner and an X-ray densitometer. Figure \ref{fig:TRACE0-BFBT} shows the void fraction measurement facility and locations. Under steady-state conditions, fine mesh void distributions were measured using the X-ray CT scanner located 50 mm above the heated length (i.e. at the assembly outlet). The X-ray densitometer measurements were performed at three different axial elevations from the bottom (i.e. 682 mm, 1706 mm and 2730 mm). For each of the four different axial locations, the cross-sectional averaged void fractions were calculated and reported in the benchmark. The steady-state void fraction data will be used in the current study, and they will be referred to from lower to upper positions as \texttt{VoidF1}, \texttt{VoidF2}, \texttt{VoidF3} and \texttt{VoidF4}, respectively.

There are 86 test cases in BFBT assembly 4101, each with a different combination of design variables including pressure, mass flow rate, power and inlet temperature. TRACE version 5.0 Patch 4 includes options for user access to 36 physical model parameters from the input file. The details of these parameters can be found in the TRACE user manual \cite{USNRC2014TRACE}. For forward UQ, the users can perturb these parameters by addition or multiplication according to expert judgment or user evaluation. In the previous inverse UQ work \cite{wu2018inversePart2}, we used global sensitivity analysis with Sobol' indices to select 5 most significant physical model parameters relevant to the BFBT benchmark. These  5 parameters are indexed as \texttt{P1008}, \texttt{P1012}, \texttt{P1022}, \texttt{P1028} and \texttt{P1029}, as shown in Table \ref{table:TRACE0-Parameters}. The nominal values are all 1.0 since they are multiplication factors. The prior ranges are chosen as $[0, 5]$ for all the parameters, which were used in design of computer experiments with LHS to generate the training data. The prior ranges are chosen to be wide to reflect the limited knowledge of these parameters. Uniform distributions are assumed for all parameters.

\begin{table}[htbp!]
	\captionsetup{justification=centering}
	\caption{Selected TRACE physical model parameters for inverse UQ.}
	\label{table:TRACE0-Parameters}
	\centering
	\begin{tabular}{l c c c}
		\toprule
		Calibration parameters $\bm{\theta}$ (multiplication factors) & Symbol & Uniform ranges  & Nominal \\ 
		\midrule
		Single phase liquid to wall heat transfer coefficient & \texttt{P1008} & (0.0, 5.0) & 1.0 \\
		Subcooled boiling heat transfer coefficient           & \texttt{P1012} & (0.0, 5.0) & 1.0 \\
		Wall drag coefficient 								  & \texttt{P1022} & (0.0, 5.0) & 1.0 \\
		Interfacial drag (bubbly/slug Rod Bundle) coefficient & \texttt{P1028} & (0.0, 5.0) & 1.0 \\
		Interfacial drag (bubbly/slug Vessel) coefficient     & \texttt{P1029} & (0.0, 5.0) & 1.0 \\
		\bottomrule
	\end{tabular}
\end{table}

To sum up, in this example, the inputs for DNNs are pressure, mass flow rate, power and inlet temperature, together with the  5 TRACE physical model parameters, \texttt{P1008}, \texttt{P1012}, \texttt{P1022}, \texttt{P1028} and \texttt{P1029}. For each of the 86 test cases in BFBT assembly 4101, 30 LHS samples for the TRACE physical model parameters are generated. In each test case, the responses are the steady-state void fraction data \texttt{VoidF1}, \texttt{VoidF2}, \texttt{VoidF3} and \texttt{VoidF4}, which will be the training outputs for the DNNs. Therefore, there are $30 \times 86 = 2580$ training data points for each void fraction response. A training dataset of half the size has been proven to be sufficient to build an accurate GP-based surrogate model in the previous work \cite{wu2018inversePart2}. In this work, a larger dataset is used since DNNs have a lot more parameters to learn.

\section{Results}
\label{section:Results}

\subsection{Results for the Bison Example}

Table \ref{table:Bison1-DNNs} lists the DNN architectures and hyperparameters used in MCD, DE and BNN for the Bison example. Because the data set is very limited in this example (200 simulations), only 5\% of the total dataset was used for validation and 10\% used for testing. In this study, we found that the DNN training process is sensitive to the hyperparameter values such as learning rate, dropout ratio, batch size, etc., hence they must be chosen with caution to have accurate uncertainty estimates. We used a grid search to select the optimum set of hyperparameters. However, this is relatively expensive in computational cost. Those hyperparameter values reported in Table \ref{table:Bison1-DNNs} are the ones based on a grid search. The numbers of epochs of training were determined based on the epochs needed when the loss function stops decreasing. One DNN is used for each PC score, so there are two DNNs in total. The reason for the output layers in DE and BNN to have 2 neurons is because NLL cost function is used. The output layers have two neurons to represent the mean values and variances of the responses.

\begin{table}[htbp!]
	\captionsetup{justification=centering}
	\caption{DNN architectures and hyperparameters used in MCD, DE and BNN for the Bison example.}
	\label{table:Bison1-DNNs}
	\centering
	\begin{tabular}{l c c c}
		\toprule
		DNN architecture/hyperparameters  &  MCD  &  DE  &  BNN \\ 
		\midrule
		number of layers                &  5  &  5  &  4  \\
		number of neurons               &  (200, 500, 500, 200, 1)  &  (50, 100, 100, 50, 2)  &  (10, 10, 10, 2)  \\
		cost/loss function              &  MSE   &  NLL  &  NLL  \\
		activation function             &  relu  & tanh  & relu  \\
		training/validation/test split  &  0.85/0.05/0.1  &  0.85/0.05/0.1  &  0.85/0.05/0.1  \\
		learning rate                   &  0.0002  &  0.0004  &  0.001  \\
		number of epochs                &  2000  &  2000  &  1000  \\
		batch size                      &  20  &  32  &  5  \\
		dropout ratio                   &  0.4  &  -  &  -  \\
		number of ensembles             &  -  &  5  &  -  \\
		number of samples for UQ        &  200  &  200  &  200  \\
		\bottomrule
	\end{tabular}
\end{table}

\begin{figure}[!ht]
	\centering
	\captionsetup{justification=centering}
	\includegraphics[width=0.95\textwidth]{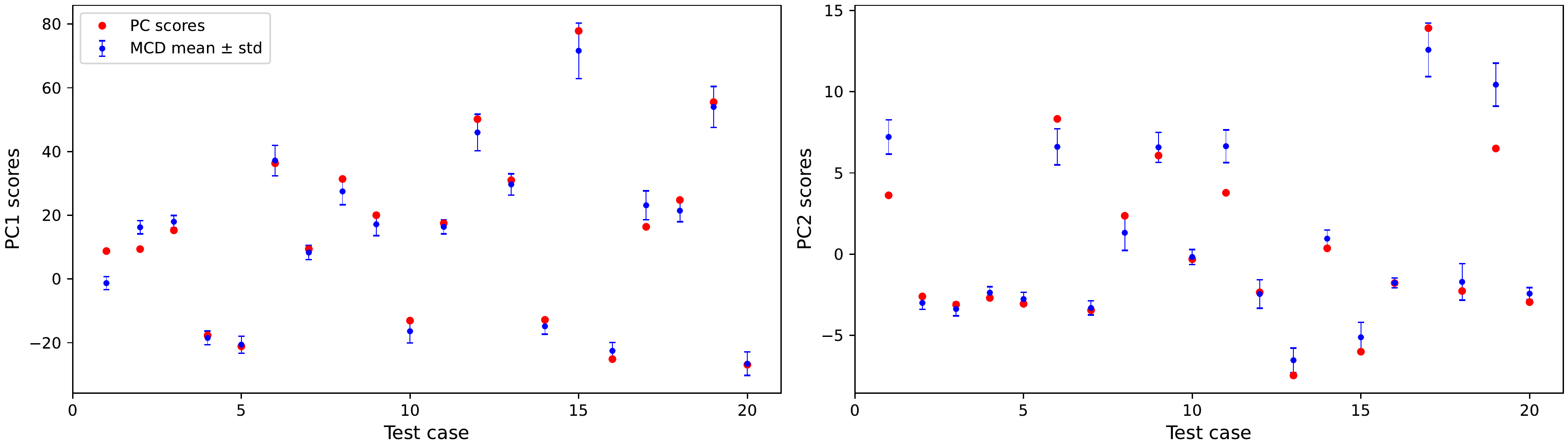}
	\caption[]{UQ of Bison FGR PC scores with MCD.}
	\label{fig:Bison1-MCD}
\end{figure}

After training the DNNs, 200 Monte Carlo simulations were performed using the DNNs in MCD and BNN to get uncertainty estimates of the testing dataset, which consists of 20 cases. For DE,  5 base learners were used. The mean and variance of the DNN results are evaluated based on Equations (\ref{eqn-DE2-DE-network-mean}) and (\ref{eqn-DE3-DE-network-variance}), respectively. Figure \ref{fig:Bison1-MCD} shows the UQ of Bison FGR PC scores using MCD, including the mean values and STDs at each test case. Because of the dropout process, generally a lot more hidden neurons are needed, as shown in Table \ref{table:Bison1-DNNs}. For PC1 scores, the DNN mean values are generally close to the test data, and the STDs are relatively small. For PC2 scores, there are a few test cases where the DNN means values are far from the test data. We have observed that the predicted PC scores from 200 Monte Carlo simulations generally follow a Gaussian distribution, so that the error bars in the figure correspond to the 68.27\% CI. If we plot the 95\% CI which is ``mean $\pm$ 2$\cdot$STD'', most of the test data points will be covered in the error bars.

\begin{figure}[!ht]
	\centering
	\captionsetup{justification=centering}
	\includegraphics[width=0.95\textwidth]{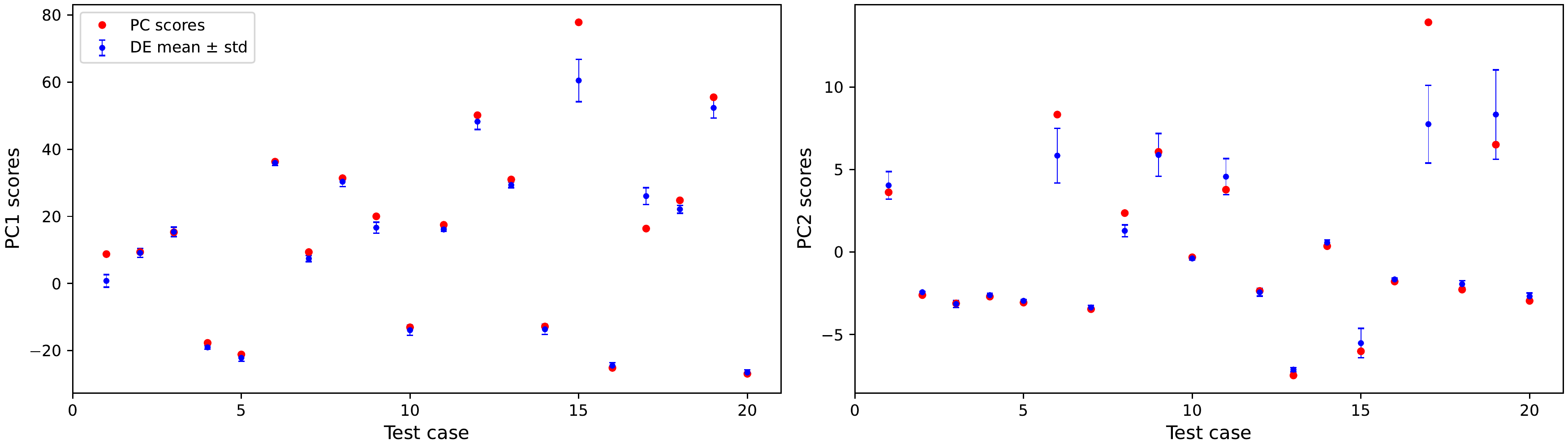}
	\caption[]{UQ of Bison FGR PC scores with DE.}
	\label{fig:Bison2-DE}
\end{figure}

Figure \ref{fig:Bison2-DE} presents the results using DE. The DNNs in DE use much smaller number of neurons compared to MCD. However, we have noticed that relatively small learning rate, large batch size and tanh activation function must be used to avoid the cost function reaching NaN values. The NaN values are caused by the nature of NLL cost function, i.e., sometimes it might take logarithm of negative values in the training process. It can be seen in Figure \ref{fig:Bison2-DE} that, for most test points, the DE means values are very close to the test data, while the uncertainty error bars are very small. There are a few test cases where the DE mean values are far from the test data, and the corresponding error bars are also wide. Similar to MCD, if 95\% CI is plotted, the test data points will all be enveloped in the error bars.

\begin{figure}[!ht]
	\centering
	\captionsetup{justification=centering}
	\includegraphics[width=0.95\textwidth]{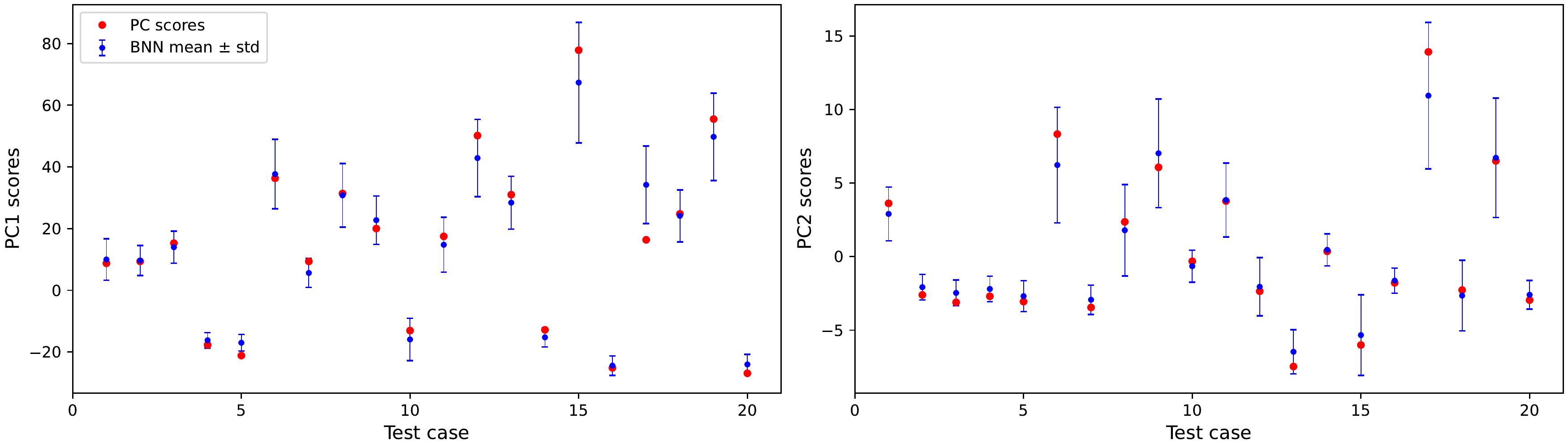}
	\caption[]{UQ of Bison FGR PC scores with BNN.}
	\label{fig:Bison3-BNN}
\end{figure}

Figure \ref{fig:Bison3-BNN} shows the results using BNN. In the training process, it was found that the neural networks must use very small number of hidden neurons, otherwise the loss function will have very large values and large oscillations. This is expected because BNN learns the means and variances of the neural network parameters (weights), making it more expensive to train. Using much smaller number of neurons is desirable. Figure \ref{fig:Bison3-BNN} shows that the means values predicted by BNNs are generally closer to the test data points, but the STDs are consistently larger, when compared to MCD and DE.

As mentioned before, the 200 samples of the PC scores at each of the 20 test cases indicate that they follow Gaussian distributions, with mean values and STDs plotted in Figures \ref{fig:Bison1-MCD} - \ref{fig:Bison3-BNN}. In order to compare the three methods w.r.t. the original FGR time series data, reverse PCA is performed to transform the PC scores to the original FGR time series. To do this, the following steps are taken:
\begin{itemize}
	\item[-] Step 1: 500 Monte Carlo samples are generated based on the Gaussian distribution at each test case, for each PC score. Note that the mean values and STDs of the Gaussian distributions are those reported in Figures \ref{fig:Bison1-MCD} - \ref{fig:Bison3-BNN}. One could have used 500 runs of the DNNs to produce the 500 samples directly. However, generating samples directly from the Gaussian distributions is more likely to have a good coverage of the distributions.
	\item[-] Step 2: The two-dimensional PC score vector at each test case is transformed to the original FGR time series using Equation (\ref{eqn-PCA1-reverse-PCA}). This is the reverse PCA process.
	\item[-] Step 3: After Step 2, at each test case, there are 500 FGR time series curves available. The mean values and STDs of these curves can be easily calculated. Then they are compared to the testing data (which is a FGR time series curve simulated by Bison) for each test case.
\end{itemize}

\begin{figure}[hbt!]
	\centering
	\captionsetup{justification=centering}
	\includegraphics[width=0.95\textwidth]{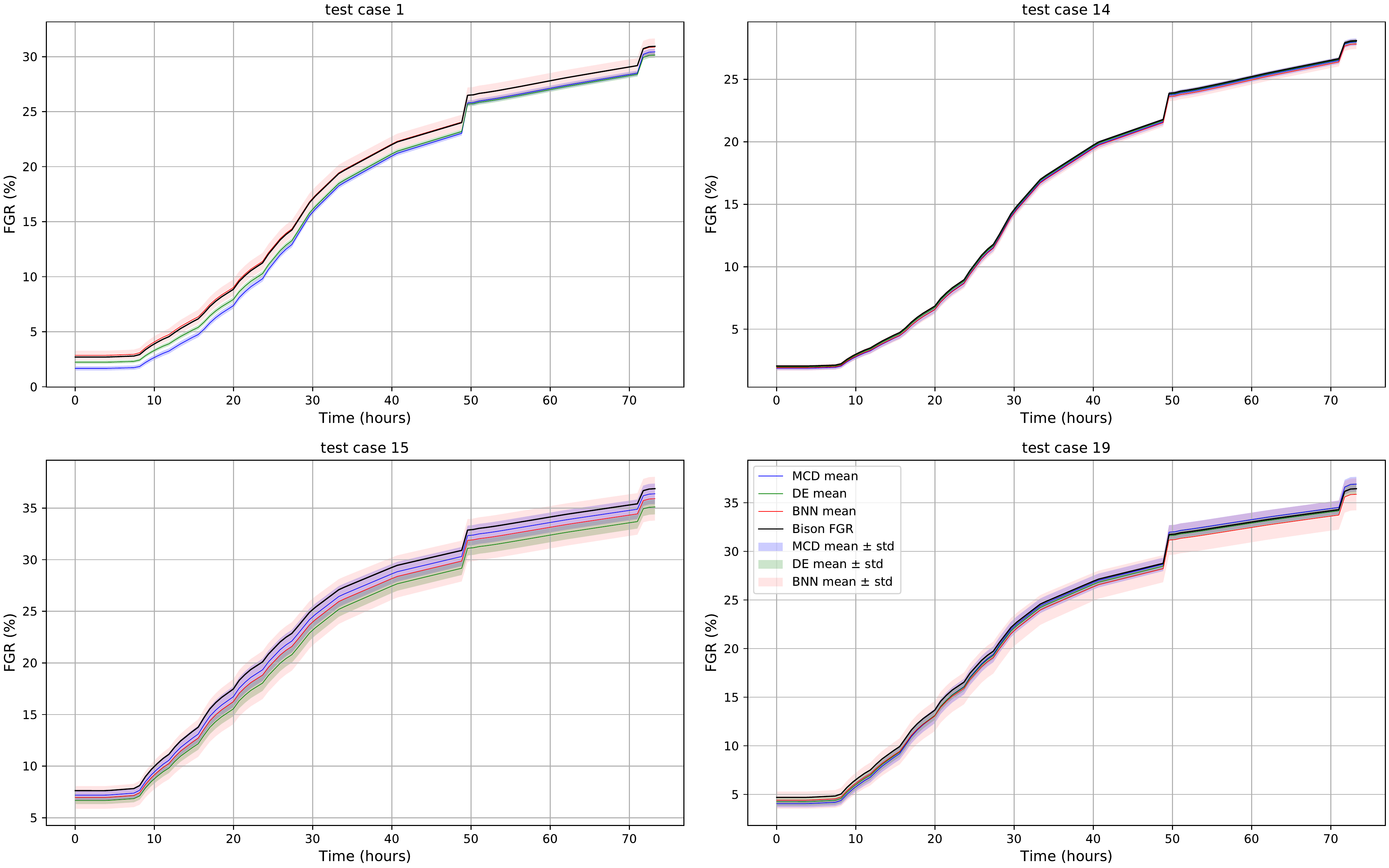}
	\caption[]{Comparison of uncertainties in the FGR obtained with MCD, DE and BNN.}
	\label{fig:Bison4-Comparison}
\end{figure}

Figure \ref{fig:Bison4-Comparison} compares the DNN UQ results with the Bison FGR simulations at four representative test cases, i.e., test case 1, 14, 15 and 19. Based on Figures \ref{fig:Bison1-MCD} - \ref{fig:Bison3-BNN}, for test case 1, BNN results of the PC scores have better agreement in the mean values, but larger STDs. This is reflected in Figure \ref{fig:Bison4-Comparison}, that the BNN mean curve agrees well with Bison simulation with relatively larger uncertainties, while MCD and DE mean curves have larger deviations from the Bison curve. In test case 14, all three methods have good mean value agreement with the PC scores and small STDs. The corresponding FGR curves comparison shows similar behavior. In test case 15, mean values from all three methods deviate from the PC scores, especially for PC1, and the STDs are larger. As a result, the mean curves deviate from the Bison curve with large uncertainties. Finally, in test case 19, the mean values from DNNs are not as far from the PC scores but the STDs are relatively large. Consequently, the mean curves from DNNs are close to the Bison curve, with larger uncertainties when compared to the test case 14.

To sum up, MCD and DE results are generally close to each other, while BNN uncertainties are consistently larger than MCD and DE. More systematic studies are needed in the future to determine if such difference is caused by the training data or the algorithms themselves. However, it can be concluded that the UQ results by MCD, DE and BNN for the PC scores in the reduced space can be consistently transformed to the original FGR time series space. However, performing UQ in the reduced PC space is more convenient since much smaller number of responses are needed.

\subsection{Results for the TRACE Example}

Table \ref{table:TRACE1-DNNs} lists the DNN architectures and hyperparameters used in MCD, DE and BNN for the TRACE example. Note that exceptions are (1) for MCD,  \texttt{VoidF1} used a learning rate of 0.001, (2) for DE,  \texttt{VoidF1} and  \texttt{VoidF2} used learning rates of 0.00025 and 0.00075, respectively, (3) for BNN,  \texttt{VoidF1} and  \texttt{VoidF2} used the tanh activation function with learning rates 0.0006 and 0.0015, respectively, while  \texttt{VoidF3} and  \texttt{VoidF4} used the relu activation function with a learning rate of 0.002. Some observations during the training process were similar to the Bison example, for example, MCD needs a lot more hidden neurons, while BNN needs the least number of hidden neurons. Also, because more training data points are available in this example, 15\% of the dataset was assigned for validation and testing, respectively. Also, it was found that the tanh activation function consistently resulted in faster convergence and smaller loss function values compared to other functions, except for  \texttt{VoidF3} and  \texttt{VoidF4} in BNN. One DNN is used for each VoidF, so there are four DNNs in total.

\begin{table}[htbp!]
	\captionsetup{justification=centering}
	\caption{DNN architectures and hyperparameters used in MCD, DE and BNN for the TRACE example.}
	\label{table:TRACE1-DNNs}
	\centering
	\begin{tabular}{l c c c}
		\toprule
		DNN architecture/hyperparameters  &  MCD  &  DE  &  BNN \\ 
		\midrule
		number of layers                &  5  &  4  &  4  \\
		number of neurons               &  (100, 200, 200, 100, 1)  &  (50, 50, 50, 2)  &  (10, 10, 10, 2)  \\
		cost/loss function              &  MSE  &  NLL  &  NLL  \\
		activation function             & tanh  & tanh  & tanh (relu)  \\
		training/validation/test split  &  0.7/0.15/0.15  &  0.7/0.15/0.15  &  0.7/0.15/0.15  \\
		learning rate                   &  0.002  &  0.001  &  0.002  \\
		number of epochs                &  2000  &  500  &  1000  \\
		batch size                      &  20  &  32  &  20  \\
		dropout ratio                   &  0.4  &  -  &  -  \\
		number of ensembles             &  -  &  5  &  -  \\
		number of samples for UQ        &  200  &  200  &  200  \\
		\bottomrule
	\end{tabular}
\end{table}

\begin{figure}[!ht]
	\centering
	\captionsetup{justification=centering}
	\includegraphics[width=0.9\textwidth]{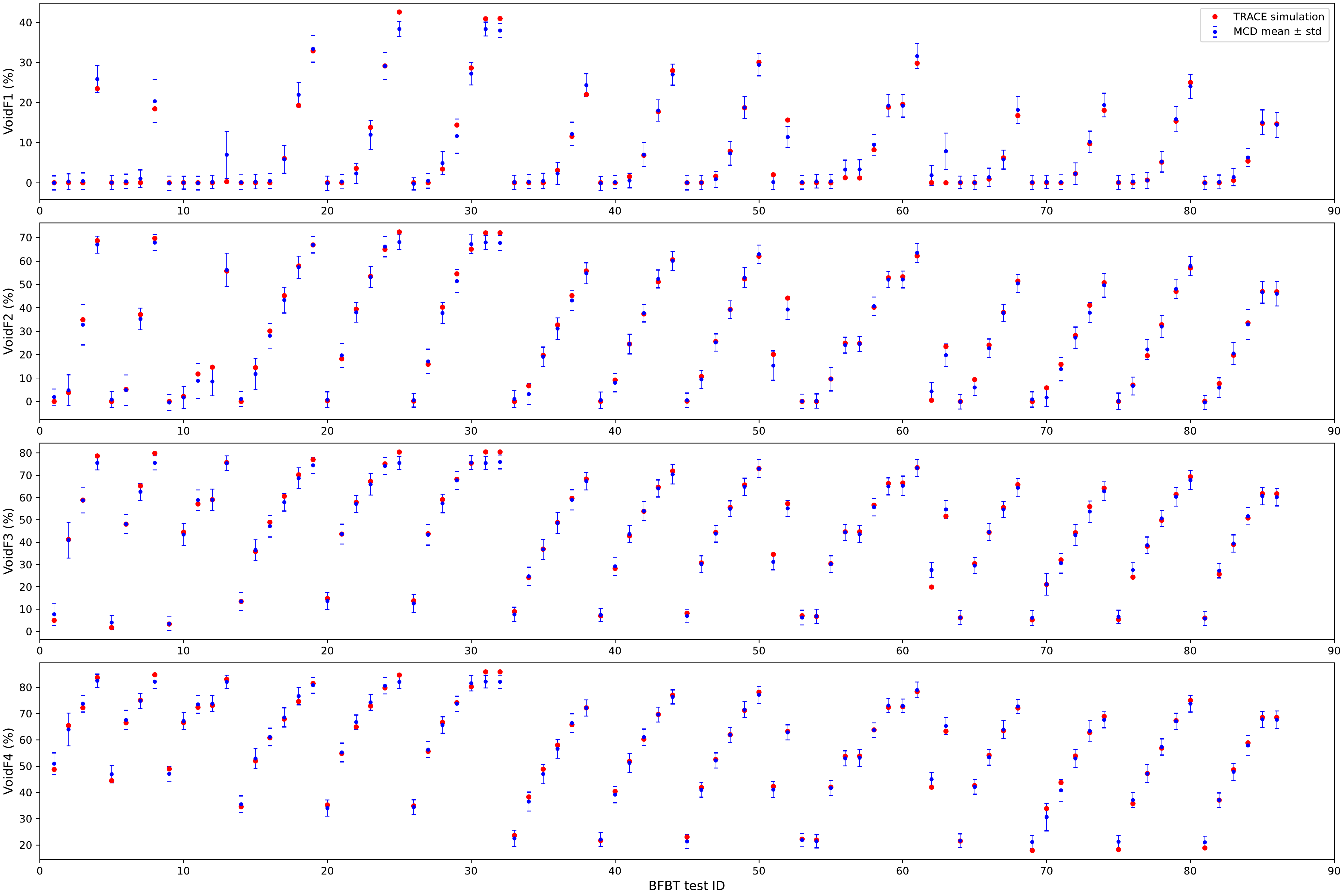}
	\caption[]{UQ of TRACE void fraction simulations with MCD.}
	\label{fig:TRACE1-MCD}
\end{figure}

\begin{figure}[!ht]
	\centering
	\captionsetup{justification=centering}
	\includegraphics[width=0.9\textwidth]{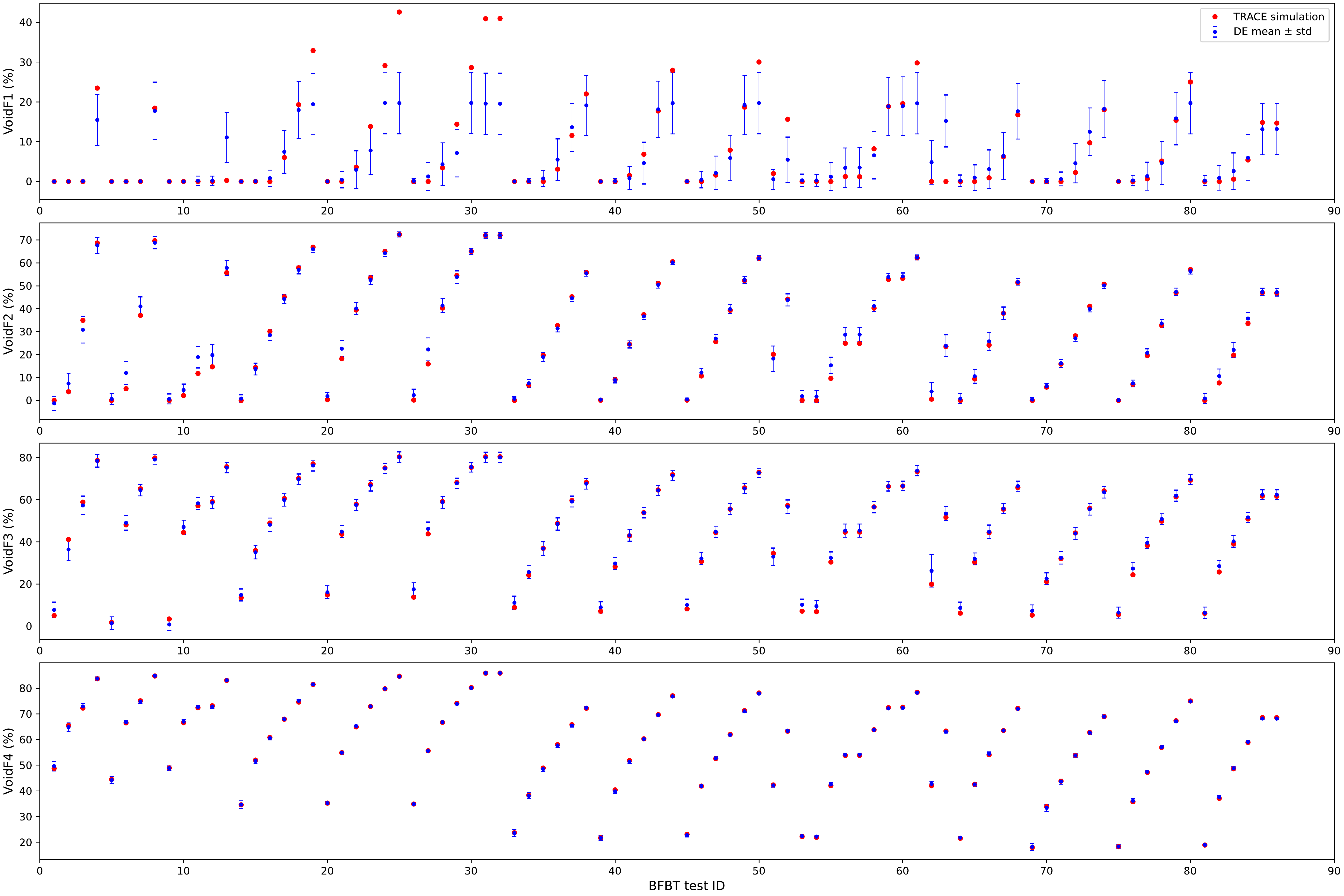}
	\caption[]{UQ of TRACE void fraction simulations with DE.}
	\label{fig:TRACE2-DE}
\end{figure}

\begin{figure}[!ht]
	\centering
	\captionsetup{justification=centering}
	\includegraphics[width=0.9\textwidth]{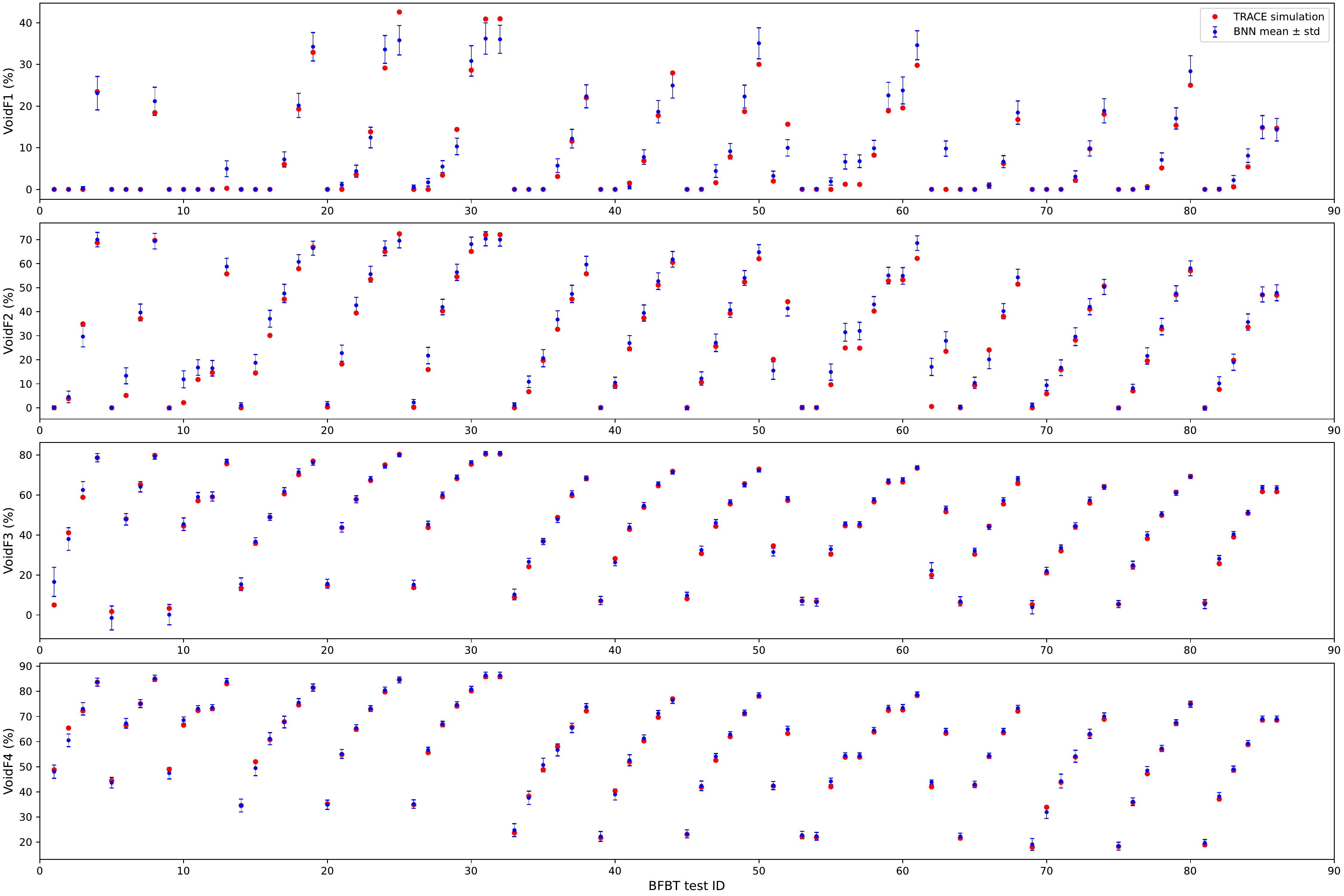}
	\caption[]{UQ of TRACE void fraction simulations with BNN.}
	\label{fig:TRACE3-BNN}
\end{figure}

Similar to the Bison example, 200 Monte Carlo samples were used for UQ of the DNNs in MCD and BNN, while Equations (\ref{eqn-DE2-DE-network-mean}) and (\ref{eqn-DE3-DE-network-variance}) were used in DE. Unlike the Bison example, in the section, we won't report the UQ results on the test dataset. Instead, TRACE simulations with nominal values (i.e., 1.0) of the 5 physical model parameters at the 86 BFBT tests will be used for the comparison. The primary reason is that these 86 test cases cover all the BFBT experimental conditions, thus providing a more complete benchmark of the DNN UQ results. Figures \ref{fig:TRACE1-MCD}, \ref{fig:TRACE2-DE} and \ref{fig:TRACE3-BNN} present the UQ results for the 4 void fraction responses using MCD, DE and BNN, respectively.

When looking at the performance of the three methods on different responses, it is obvious that the UQ results on  \texttt{VoidF1} is the  worst, especially for DE and BNN. The mean values of DNNs overall have larger deviations with the TRACE simulation data, while the STDs are much larger than the other responses. The main reason is that,  \texttt{VoidF1} is not as ``well-structured'' than  \texttt{VoidF2} -  \texttt{VoidF4}. This is because for many BFBT cases in which the power is low, inlet temperature is low, pressure is high and mass flow rate is high, the phase change will be slower at the bottom locations of the bundles. As a result,  \texttt{VoidF1} will be 0.0\%, as shown by many points in Figures \ref{fig:TRACE1-MCD} - \ref{fig:TRACE3-BNN}. In other words, in the training dataset, many samples with different combinations of the 5 physical model parameters all have the same 0.0\% training output for  \texttt{VoidF1}. Such ``many-to-one'' function mapping is more difficult to fit by ML regression algorithms. It was observed that MCD is the most robust method for  \texttt{VoidF1} while DE has the worst performance for  \texttt{VoidF1}. DE can predict very well when \texttt{VoidF1} equals 0.0\% with tiny STDs. But when  \texttt{VoidF1} is above 10.0\%, the mean values differ from TRACE results and the STDs are very large.

For  \texttt{VoidF2} -  \texttt{VoidF4}, all three methods have very accurate mean values when compared to TRACE simulation results. MCD generally produces small and similar STDs for all the responses in all the test cases. For DE and BNN, it can be observed that overall the STDs are small when the differences between DNN prediction and TRACE simulation are small. For  \texttt{VoidF4}, DE and BNN produce tiny STDs, because unlike  \texttt{VoidF1},  \texttt{VoidF4} is better structured (no ``many-to-one'' input-output mapping). It can be concluded that the performance of these three DNN UQ methods, especially DE and BNN, depends on the quality of the dataset. Therefore, the users should use great caution when applying these methods for UQ.

The quality of prediction uncertainties obtained using BNNs crucially depends on (i) the degree of approximation due to computational constraints and (ii) whether the prior distribution is reasonable. In practice, BNNs are often harder to implement and computationally slower to train compared to non-BNNs. This is the primary reason for us to use very small number of hidden neurons for BNNs and non-informative priors such as wide uniform distributions. Nevertheless, BNNs with simple architectures still produced satisfactory UQ results in both examples. Furthermore, in this work, different DNN architectures (number of hidden layers and neurons) were used in different methods. They were selected from a hyperparameter optimization process by grid search. Once might argue that in order to have a ``fair'' comparison, the same DNN architecture and hyperparameters should be used for these three methods. While this makes intuitive sense, we argue that MCD, DE and BNN are very different methods by design, using the same DNN architecture and hyperparameters may cause problems. For example, MCD should always use larger DNNs (more layers and/or hidden neurons) because a large amount of the hidden neurons will be dropped randomly in the training process. If the DNN is too shallow and/or narrow, the expressive power after dropout will be low. On the other hand, BNN prefers smaller number of layers and/or hidden neurons, because it learns the distributions of the DNN parameters. The computational cost of VI will increase quickly with more hidden neurons. It has also been noticed that, MCD is generally more robust to train than DE and BNN. Because DE and BNN use the NLL cost function that has a logarithm operation. However, in the two examples we have noticed that NLL cost function converges faster than the MSE cost function.

\section{Conclusion}
\label{section:Conclusions}

In this work, we presented results from a preliminary investigation in quantification of prediction or approximation uncertainties of DNNs when used as surrogate models for Bison and TRACE codes. UQ of DNNs, or any ML models, is important to establish confidence in DNN predictions, especially when they are generalized for extrapolated domains. Three methods, MCD, DE and BNN, are compared based on two demonstration problems, Bison simulations of time-dependent fission gas release and TRACE simulations of BFBT void fraction data. It was found that the three methods typically require different DNN architectures and hyperparameters to optimize their performance. The UQ results also depend on the amount of training data available and the nature of the data. Overall, all these three methods can provide reasonable estimations of the approximation uncertainties. The uncertainties are generally smaller when the mean predictions are close to the test data, but the BNN methods usually produce larger uncertainties than MCD and DE. Future work will be focused on using more complicated problems to determine whether such disagreement in quantified uncertainties is caused by algorithms or the training data.

\pagebreak
\bibliographystyle{style/ans_js}
\bibliography{bibliography.bib}

\begin{thebibliography}{10}
\newcommand{\enquote}[1]{``#1''}
\providecommand{\url}[1]{\texttt{#1}}
\providecommand{\urlprefix}{URL }
\expandafter\ifx\csname urlstyle\endcsname\relax
  \providecommand{\doi}[1]{doi:\discretionary{}{}{}#1}\else
  \providecommand{\doi}{doi:\discretionary{}{}{}\begingroup
  \urlstyle{rm}\Url}\fi

\bibitem{baker2019workshop}
\textsc{N.~Baker}, \textsc{F.~Alexander}, \textsc{T.~Bremer},
  \textsc{A.~Hagberg}, \textsc{Y.~Kevrekidis}, \textsc{H.~Najm},
  \textsc{M.~Parashar}, \textsc{A.~Patra}, \textsc{J.~Sethian},
  \textsc{S.~Wild} \textsc{et~al.}, \enquote{Workshop report on basic research
  needs for scientific machine learning: Core technologies for artificial
  intelligence,} , USDOE Office of Science (SC), Washington, DC (United States)
  (2019).

\bibitem{solomatine2009data}
\textsc{D.~Solomatine}, \textsc{L.~M. See}, and \textsc{R.~Abrahart},
  \enquote{Data-driven modelling: concepts, approaches and experiences,}
  \emph{Practical hydroinformatics}, 17--30 (2009).

\bibitem{kapteyn2020physics}
\textsc{M.~G. Kapteyn} and \textsc{K.~E. Willcox}, \enquote{From physics-based
  models to predictive digital twins via interpretable machine learning,}
  \emph{arXiv preprint arXiv:2004.11356} (2020).

\bibitem{liu2019validation}
\textsc{Y.~Liu} and \textsc{N.~Dinh}, \enquote{Validation and uncertainty
  quantification for wall boiling closure relations in multiphase-CFD solver,}
  \emph{Nuclear Science and Engineering}, \textbf{193}, \emph{1-2}, 81 (2019).

\bibitem{zhao2020prediction}
\textsc{X.~Zhao}, \textsc{K.~Shirvan}, \textsc{R.~K. Salko}, and
  \textsc{F.~Guo}, \enquote{On the prediction of critical heat flux using a
  physics-informed machine learning-aided framework,} \emph{Applied Thermal
  Engineering}, \textbf{164}, 114540 (2020).

\bibitem{bao2021deep}
\textsc{H.~Bao}, \textsc{J.~Feng}, \textsc{N.~Dinh}, and \textsc{H.~Zhang},
  \enquote{Deep learning interfacial momentum closures in coarse-mesh CFD
  two-phase flow simulation using validation data,} \emph{International Journal
  of Multiphase Flow}, \textbf{135}, 103489 (2021).

\bibitem{aguiar2020bringing}
\textsc{J.~A. Aguiar}, \textsc{A.~M. Jokisaari}, \textsc{M.~Kerr}, and
  \textsc{R.~Allen~Roach}, \enquote{Bringing nuclear materials discovery and
  qualification into the 21st century,} \emph{Nature Communications},
  \textbf{11}, \emph{1}, 1 (2020).

\bibitem{lee2021development}
\textsc{J.~Lee}, \textsc{L.~Lin}, \textsc{P.~Athe}, and \textsc{N.~Dinh},
  \enquote{Development of the machine learning-based safety significant factor
  inference model for diagnosis in autonomous control system,} \emph{Annals of
  Nuclear Energy}, \textbf{162}, 108443 (2021).

\bibitem{lin2021development}
\textsc{L.~Lin}, \textsc{P.~Athe}, \textsc{P.~Rouxelin}, \textsc{M.~Avramova},
  \textsc{A.~Gupta}, \textsc{R.~Youngblood}, \textsc{J.~Lane}, and
  \textsc{N.~Dinh}, \enquote{Development and assessment of a nearly autonomous
  management and control system for advanced reactors,} \emph{Annals of Nuclear
  Energy}, \textbf{150}, 107861 (2021).

\bibitem{lin2022digital}
\textsc{L.~Lin}, \textsc{P.~Athe}, \textsc{P.~Rouxelin}, \textsc{M.~Avramova},
  \textsc{A.~Gupta}, \textsc{R.~Youngblood}, \textsc{J.~Lane}, and
  \textsc{N.~Dinh}, \enquote{Digital-twin-based improvements to diagnosis,
  prognosis, strategy assessment, and discrepancy checking in a nearly
  autonomous management and control system,} \emph{Annals of Nuclear Energy},
  \textbf{166}, 108715 (2022).

\bibitem{vicente2021nuclear}
\textsc{P.~Vicente-Valdez}, \textsc{L.~Bernstein}, and \textsc{M.~Fratoni},
  \enquote{Nuclear data evaluation augmented by machine learning,} \emph{Annals
  of Nuclear Energy}, \textbf{163}, 108596 (2021).

\bibitem{shriver2021prediction}
\textsc{F.~Shriver}, \textsc{C.~Gentry}, and \textsc{J.~Watson},
  \enquote{Prediction of neutronics parameters within a two-dimensional
  reflective PWR assembly using deep learning,} \emph{Nuclear Science and
  Engineering}, \textbf{195}, \emph{6}, 626 (2021).

\bibitem{wu2018inversePart1}
\textsc{X.~Wu}, \textsc{T.~Kozlowski}, \textsc{H.~Meidani}, and
  \textsc{K.~Shirvan}, \enquote{Inverse uncertainty quantification using the
  modular Bayesian approach based on Gaussian process, Part 1: Theory,}
  \emph{Nuclear Engineering and Design}, \textbf{335}, 339 (2018).

\bibitem{wu2018inversePart2}
\textsc{X.~Wu}, \textsc{T.~Kozlowski}, \textsc{H.~Meidani}, and
  \textsc{K.~Shirvan}, \enquote{Inverse uncertainty quantification using the
  modular Bayesian approach based on Gaussian Process, Part 2: Application to
  TRACE,} \emph{Nuclear Engineering and Design}, \textbf{335}, 417 (2018).

\bibitem{xie2021towards}
\textsc{Z.~Xie}, \textsc{F.~Alsafadi}, and \textsc{X.~Wu}, \enquote{Towards
  improving the predictive capability of computer simulations by integrating
  inverse Uncertainty Quantification and quantitative validation with Bayesian
  hypothesis testing,} \emph{Nuclear Engineering and Design}, \textbf{383},
  111423 (2021).

\bibitem{moloko2022quantification}
\textsc{L.~Moloko}, \textsc{P.~Bokov}, \textsc{X.~Wu}, and \textsc{K.~Ivanov},
  \enquote{Quantification of Neural Networks Uncertainties with Applications to
  SAFARI-1 Axial Neutron Flux Profiles,} \emph{Proceedings of the International
  Conference on Physics of Reactors (PHYSOR) 2022}, Pittsburgh, PA, USA, May
  15–20, 2022. (2022).

\bibitem{oberkampf2010verification}
\textsc{W.~L. Oberkampf} and \textsc{C.~J. Roy}, \emph{Verification and
  validation in scientific computing}, Cambridge University Press (2010).

\bibitem{abdar2021review}
\textsc{M.~Abdar}, \textsc{F.~Pourpanah}, \textsc{S.~Hussain},
  \textsc{D.~Rezazadegan}, \textsc{L.~Liu}, \textsc{M.~Ghavamzadeh},
  \textsc{P.~Fieguth}, \textsc{X.~Cao}, \textsc{A.~Khosravi}, \textsc{U.~R.
  Acharya} \textsc{et~al.}, \enquote{A review of uncertainty quantification in
  deep learning: Techniques, applications and challenges,} \emph{Information
  Fusion} (2021).

\bibitem{psaros2022uncertainty}
\textsc{A.~F. Psaros}, \textsc{X.~Meng}, \textsc{Z.~Zou}, \textsc{L.~Guo}, and
  \textsc{G.~E. Karniadakis}, \enquote{Uncertainty Quantification in Scientific
  Machine Learning: Methods, Metrics, and Comparisons,} \emph{arXiv preprint
  arXiv:2201.07766} (2022).

\bibitem{wu2021comprehensive}
\textsc{X.~Wu}, \textsc{Z.~Xie}, \textsc{F.~Alsafadi}, and
  \textsc{T.~Kozlowski}, \enquote{A comprehensive survey of inverse uncertainty
  quantification of physical model parameters in nuclear system
  thermal--hydraulics codes,} \emph{Nuclear Engineering and Design},
  \textbf{384}, 111460 (2021).

\bibitem{gal2016dropout}
\textsc{Y.~Gal} and \textsc{Z.~Ghahramani}, \enquote{Dropout as a bayesian
  approximation: Representing model uncertainty in deep learning,}
  \emph{international conference on machine learning}, 1050--1059, PMLR (2016).

\bibitem{lakshminarayanan2017simple}
\textsc{B.~Lakshminarayanan}, \textsc{A.~Pritzel}, and \textsc{C.~Blundell},
  \enquote{Simple and scalable predictive uncertainty estimation using deep
  ensembles,} \emph{Advances in neural information processing systems},
  \textbf{30} (2017).

\bibitem{blundell2015weight}
\textsc{C.~Blundell}, \textsc{J.~Cornebise}, \textsc{K.~Kavukcuoglu}, and
  \textsc{D.~Wierstra}, \enquote{Weight uncertainty in neural network,}
  \emph{International Conference on Machine Learning}, 1613--1622, PMLR (2015).

\bibitem{pastore2013physics}
\textsc{G.~Pastore}, \textsc{L.~Luzzi}, \textsc{V.~Di~Marcello}, and
  \textsc{P.~Van~Uffelen}, \enquote{Physics-based modelling of fission gas
  swelling and release in UO2 applied to integral fuel rod analysis,}
  \emph{Nuclear Engineering and Design}, \textbf{256}, 75 (2013).

\bibitem{williamson2012multidimensional}
\textsc{R.~Williamson}, \textsc{J.~Hales}, \textsc{S.~Novascone},
  \textsc{M.~Tonks}, \textsc{D.~Gaston}, \textsc{C.~Permann},
  \textsc{D.~Andrs}, and \textsc{R.~Martineau}, \enquote{Multidimensional
  multiphysics simulation of nuclear fuel behavior,} \emph{Journal of Nuclear
  Materials}, \textbf{423}, \emph{1}, 149 (2012).

\bibitem{killeen2006fuel}
\textsc{J.~Killeen}, \textsc{J.~Turnbull}, and \textsc{E.~Sartori},
  \enquote{Fuel modelling at extended burnup: IAEA coordinated research project
  FUMEX-II,} \emph{Transactions of the Top Fuel 2006 International Meeting on
  LWR Fuel Performance}, 22--26 (2006).

\bibitem{USNRC2014TRACE}
\textsc{USNRC}, \emph{TRAC/RELAP Advanced Computational Engine (TRACE) V5.840
  User's Manual, Volume 1: Input Specification}, Division of Safety Analysis,
  Office of Nuclear Regulatory Research, U. S. Nuclear Regulatory Commission,
  Washington, DC. (2014).

\bibitem{neykov2005nupec}
\textsc{B.~Neykov}, \textsc{F.~Aydogan}, \textsc{L.~Hochreiter},
  \textsc{K.~Ivanov}, \textsc{H.~Utsuno}, \textsc{F.~Kasahara},
  \textsc{E.~Sartori}, and \textsc{M.~Martin}, \emph{NUPEC BWR full-size
  fine-mesh bundle test (BFBT) benchmark}, OECD/NEA, NEA/NSC/DOC(2005)5 (2005).

\bibitem{srivastava2014dropout}
\textsc{N.~Srivastava}, \textsc{G.~Hinton}, \textsc{A.~Krizhevsky},
  \textsc{I.~Sutskever}, and \textsc{R.~Salakhutdinov}, \enquote{Dropout: a
  simple way to prevent neural networks from overfitting,} \emph{The journal of
  machine learning research}, \textbf{15}, \emph{1}, 1929 (2014).

\bibitem{ghahramani2016history}
\textsc{Z.~Ghahramani}, \enquote{A history of bayesian neural networks,}
  \emph{NIPS workshop on Bayesian deep learning} (2016).

\bibitem{goan2020bayesian}
\textsc{E.~Goan} and \textsc{C.~Fookes}, \enquote{Bayesian neural networks: An
  introduction and survey,} \emph{Case Studies in Applied Bayesian Data
  Science}, 45--87, Springer, Cham.

\bibitem{neal2012bayesian}
\textsc{R.~M. Neal}, \emph{Bayesian learning for neural networks}, vol. 118,
  Springer Science \& Business Media (2012).

\bibitem{blei2017variational}
\textsc{D.~M. Blei}, \textsc{A.~Kucukelbir}, and \textsc{J.~D. McAuliffe},
  \enquote{Variational inference: A review for statisticians,} \emph{Journal of
  the American statistical Association}, \textbf{112}, \emph{518}, 859 (2017).

\bibitem{tzikas2008variational}
\textsc{D.~G. Tzikas}, \textsc{A.~C. Likas}, and \textsc{N.~P. Galatsanos},
  \enquote{The variational approximation for Bayesian inference,} \emph{IEEE
  Signal Processing Magazine}, \textbf{25}, \emph{6}, 131 (2008).

\bibitem{pastore2015uncertainty}
\textsc{G.~Pastore}, \textsc{L.~Swiler}, \textsc{J.~D. Hales}, \textsc{S.~R.
  Novascone}, \textsc{D.~M. Perez}, \textsc{B.~W. Spencer}, \textsc{L.~Luzzi},
  \textsc{P.~Van~Uffelen}, and \textsc{R.~L. Williamson}, \enquote{Uncertainty
  and sensitivity analysis of fission gas behavior in engineering-scale fuel
  modeling,} \emph{Journal of Nuclear Materials}, \textbf{456}, 398 (2015).

\bibitem{wu2018kriging}
\textsc{X.~Wu}, \textsc{T.~Kozlowski}, and \textsc{H.~Meidani},
  \enquote{Kriging-based inverse uncertainty quantification of nuclear fuel
  performance code BISON fission gas release model using time series
  measurement data,} \emph{Reliability Engineering \& System Safety},
  \textbf{169}, 422 (2018).

\bibitem{williamson2016validating}
\textsc{R.~Williamson}, \textsc{K.~Gamble}, \textsc{D.~Perez},
  \textsc{S.~Novascone}, \textsc{G.~Pastore}, \textsc{R.~Gardner},
  \textsc{J.~Hales}, \textsc{W.~Liu}, and \textsc{A.~Mai}, \enquote{Validating
  the BISON fuel performance code to integral LWR experiments,} \emph{Nuclear
  Engineering and Design}, \textbf{301}, 232 (2016).

\bibitem{shlens2014tutorial}
\textsc{J.~Shlens}, \enquote{A tutorial on principal component analysis,}
  \emph{arXiv preprint arXiv:1404.1100} (2014).

\end{thebibliography}

\end{document}